\pdfoutput=1
\documentclass[10pt,twocolumn,letterpaper]{article}

\usepackage[pagenumbers]{cvpr}
\usepackage{cuted}

\usepackage{currfile}

\usepackage{caption}

\usepackage{microtype}

\renewcommand{\paragraph}[1]{\vspace{.45em}\noindent\textbf{#1}\hspace{.6em}}

\usepackage{bm}
\usepackage{multirow}
\usepackage{colortbl}
\usepackage{xspace}

\newcommand{\SE}{\mathrm{SE}(3)}
\newcommand{\Sim}{\mathrm{Sim}(3)}

\DeclareMathOperator*{\argmin}{arg\,min}

\newcommand{\method}{Accord\xspace}

\newcommand{\na}{--}

\usepackage{tikz}
\usetikzlibrary{positioning,arrows.meta,fit,backgrounds,calc}
\definecolor{teach}{RGB}{200,80,30}
\definecolor{basegray}{RGB}{110,110,110}

\newcommand{\resFusionOriEuroc}{0.375}
\newcommand{\resFusionOriHandheld}{0.283}
\newcommand{\resInputBaseColour}{0.671}
\newcommand{\resInputBaseGrey}{0.672}
\newcommand{\resInputGiantColour}{0.699}
\newcommand{\resInputGiantGrey}{0.781}
\newcommand{\resLiveBaseGpu}{2.5}
\newcommand{\resLiveBaseRFour}{0.602}
\newcommand{\resLiveGiantAteFinalOri}{0.072}
\newcommand{\resLiveGiantAteOdometryOri}{0.142}
\newcommand{\resLiveGiantGpu}{8.9}
\newcommand{\resLiveLargeOriSpread}{0.031}
\newcommand{\resLiveLargeRFour}{0.604}
\newcommand{\resOriN}{5}
\newcommand{\resOtherFusionRFour}{0.599}
\newcommand{\resRoomsN}{6}

\newcommand{\resHiltiElevenFinal}{0.160}
\newcommand{\resHiltiElevenOdometry}{0.122}
\newcommand{\resHiltiFourFinal}{0.281}
\newcommand{\resHiltiFusionChoiceEuroc}{0.067}
\newcommand{\resHiltiFusionChoiceHandheld}{0.364}
\newcommand{\resHiltiFusionChosen}{hand-held}
\newcommand{\resHiltiOneFinal}{0.544}

\newcommand{\ovEurocTrajOrb}{3.2}
\newcommand{\ovEurocTrajOurs}{6.8}
\newcommand{\ovTumviTrajOkvis}{1.4}
\newcommand{\ovTumviTrajOurs}{4.5}
\newcommand{\ovHiltiTrajOkvis}{17.1}
\newcommand{\ovHiltiTrajOurs}{37.0}
\newcommand{\ovSpiresTrajOurs}{73.0}
\newcommand{\ovSpiresTrajOursLive}{61.0}
\newcommand{\setSeqN}{30}
\newcommand{\setMinutes}{91}
\newcommand{\setKm}{4.3}

\newcommand{\armRecOriCorr}{0.782}
\newcommand{\armRecRoomsCorr}{0.570}
\newcommand{\armRefOriCorr}{0.891}
\newcommand{\armRefRoomsCorr}{0.614}
\newcommand{\armScarfGiantOri}{0.742}
\newcommand{\armScarfGiantRooms}{0.203}
\newcommand{\armScarfOwnOri}{0.742}
\newcommand{\armScarfOwnRooms}{0.205}
\newcommand{\cumAnchorHiltiCorr}{0.147}
\newcommand{\cumAnchorRoomsCorr}{0.547}
\newcommand{\cumNoneHiltiCorr}{0.120}
\newcommand{\cumNoneRoomsCorr}{0.488}
\newcommand{\cumSystemAheadHilti}{2}
\newcommand{\cumSystemAheadRooms}{6}
\newcommand{\cumSystemHiltiCorr}{0.164}
\newcommand{\cumSystemRoomsCorr}{0.570}
\newcommand{\fvioEurocAteMaxKm}{22.3}
\newcommand{\fvioEurocAteMinKm}{2.2}
\newcommand{\fvioEurocDiverged}{6}
\newcommand{\fvioEurocRooms}{6}
\newcommand{\fvioOriAheadOurs}{5}
\newcommand{\fvioOriMap}{0.556}
\newcommand{\fvioScarfPaper}{73.1}
\newcommand{\fvioScarfRunCut}{64.5}
\newcommand{\nzEurocGap}{-0.043}
\newcommand{\nzEurocHi}{0.589}
\newcommand{\nzEurocLo}{0.521}
\newcommand{\nzOriGap}{+0.005}
\newcommand{\nzOriHi}{0.803}
\newcommand{\nzOriLo}{0.782}
\newcommand{\oursScarfCut}{74.3}
\newcommand{\oursScarfCutAheadPaper}{4}
\newcommand{\oursScarfCutAheadRun}{5}
\newcommand{\ovEurocTrajFilter}{7.6}
\newcommand{\ovHiltiTrajFilter}{49.0}
\newcommand{\ovHiltiTrajOkvisLive}{27.4}
\newcommand{\ovTumviTrajFilter}{6.8}
\newcommand{\ovTumviTrajOkvisLive}{7.3}
\newcommand{\pobDrainHi}{40}
\newcommand{\pobDrainLo}{13}
\newcommand{\spAheadMap}{3}
\newcommand{\spAheadTraj}{2}
\newcommand{\spFvioMap}{0.116}
\newcommand{\spFvioTraj}{1.05}
\newcommand{\spOursMap}{0.196}
\newcommand{\spOursTraj}{0.73}
\newcommand{\stOriFilter}{-1.3}
\newcommand{\stRoomsFilter}{-0.4}

\newcommand{\phAndroidLead}{6}
\newcommand{\phAndroidN}{10}
\newcommand{\phIphoneLead}{4}
\newcommand{\phIphoneN}{11}
\newcommand{\phAndroidCommonFvioOurs}{0.94}
\newcommand{\phAndroidCommonFvioOther}{1.99}
\newcommand{\phIphoneCommonFvioOurs}{1.06}
\newcommand{\phIphoneCommonFvioOther}{1.54}
\newcommand{\phWalksN}{11}
\newcommand{\phWalksWord}{eleven}
\newcommand{\phFirstN}{9}
\newcommand{\phMinutes}{59}
\newcommand{\phKm}{2.5}
\newcommand{\phIndoorN}{2}
\newcommand{\phOutdoorN}{7}
\newcommand{\phNightN}{2}
\newcommand{\phIphoneBelowCm}{13.7}
\newcommand{\phFailM}{10}
\newcommand{\phAndroidImuHz}{210}
\newcommand{\phIphoneImuHz}{100}
\newcommand{\phAndroidCamHz}{30}
\newcommand{\phIphoneCamHzMin}{19}
\newcommand{\phIphoneCamHzMax}{30}
\newcommand{\phGyroCorrMin}{0.998}

\newcommand{\gpMinutes}{11.7}
\newcommand{\gpMetres}{623}
\newcommand{\gpFrameHz}{30}
\newcommand{\gpImuHz}{198}

\newcommand{\specSeqN}{17}
\newcommand{\specOursLead}{12}
\newcommand{\specOursTopTwo}{16}

\newcommand{\corrAbsMedianRooms}{1.6}
\newcommand{\corrAbsMedianOri}{1.5}

\usepackage{float}
\usepackage{placeins}
\definecolor{cvprblue}{rgb}{0.21,0.49,0.74}
\usepackage[pagebackref,breaklinks,colorlinks,allcolors=cvprblue]{hyperref}

\title{Accord: Universal Online Metric Localization and Dense Mapping \\ from one Camera and an IMU}

\newcommand{\pdfauthors}{Jaafar Mahmoud, Arthur Movsesyan, Cuong Huynh, Mikhail Iumanov, Sergey Zagoruyko, Sergey Kolyubin}
\author{Jaafar Mahmoud \qquad Arthur Movsesyan \qquad Cuong Huynh \qquad Mikhail Iumanov \\ \qquad Sergey Zagoruyko \qquad Sergey Kolyubin\\[3pt]
{\normalsize Biomechatronics and Energy-Efficient Robotics (BE2R) Lab, ITMO University, Saint Petersburg, Russia}\\[2pt]
{\normalsize \texttt{jaafar.a.mahmoud@itmo.ru} \qquad Code: \url{https://github.com/be2rlab/ACCORD}}}

\hypersetup{pdftitle={Accord: Universal Online Metric Localization and Dense Mapping from one Camera and an IMU},
  pdfauthor={\pdfauthors}}

\begin{document}
\twocolumn[{%
\renewcommand\twocolumn[1][]{#1}%
\maketitle
\vspace{-1.6em}
\begin{center}
  \includegraphics[width=0.85\textwidth]{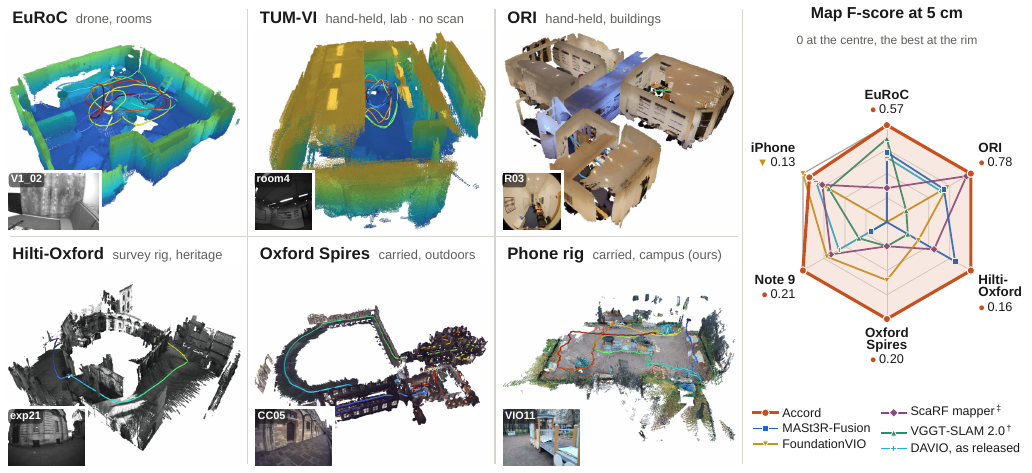}
  \captionof{figure}{\textbf{One configuration, six platforms.} The dense metric map \method builds online from one camera and an IMU, with the same settings everywhere, beside a frame of its camera, each map placed by the system's own trajectory.}
  \label{fig:teaser}
\end{center}

\vspace{.8em}
}]

\begin{abstract}
Dense metric mapping from a single camera and IMU remains challenging: visual--inertial odometry (VIO) provides metric motion but sparse geometry, whereas feed-forward depth models predict dense structure without ensuring metric scale. We present \method, which constructs \textbf{an accurate online dense metric map from any camera with an IMU}. A VIO filter publishes metric poses, a depth foundation model conditioned on them predicts depth for short keyframe windows, and a factor graph fuses the resulting submaps. Recent work has explored several approaches to this problem. What is essential is the multi-view agreement (\method), which is considered in SLAM as a source of geometric self-consistency. \method anchors each keyframe window to metric 3D points triangulated from the filter poses, then exploits bidirectional dense depth agreement between co-visible images to estimate and correct per-image scale errors while preserving the metric scale established by the anchors. Context keyframes further improve depth prediction, and the resulting submaps are fused in a factor graph. We evaluate a single mapping configuration across five public camera--IMU setups and a new Phone-rig/GoPro dataset. With nothing tuned per dataset/setup and every map aligned only by \method's own trajectory, \method achieves consistently best/second-best mean F-score at 5 cm on every platform. \method is open-source along with the phone-rig dataset for the benefit of the research community.
\end{abstract}

\section{Introduction}\label{sec:intro}
A robot that moves and acts autonomously plans in a dense, metric map of the space its camera moves through. Yet dense metric mapping and localization still rely on depth cameras, LiDAR, or specific custom sensor rigs, even though nearly every handheld device (smartphones included) already carries a camera and an inertial measurement unit (IMU). Two lines of works each recover part of what is needed from these sensors. Visual--inertial odometry (VIO) estimates metric motion in real-time~\cite{mourikis2007msckf,geneva2020openvins,qin2018vinsmono,campos2021orbslam3}, but its map consists of only a few hundred sparse landmarks. On the other hand, feed-forward geometry models~\cite{wang2024dust3r,leroy2024mast3r,wang2025vggt,wang2025pi3,keetha2026mapanything,lin2025da3} predict dense structure from uncalibrated images in a single forward pass, but not its metric scale and have no sense of gravity direction. SLAM systems built on them align their submaps only up to a similarity or projective transformation~\cite{murai2025mast3rslam,maggio2026vggtslam2}, and are evaluated against the reference only after such an alignment.

The two are complementary, and recent work combines learned regressed geometry with visual--inertial estimation~\cite{zhang2026ffvioinit,zhang2026scarfslam,salih2026vidar,zhou2026mast3rfusion,rhee2026foundationvio,davio2026}. We build on DAVIO~\cite{davio2026}, which conditions a depth foundation model, Depth Anything 3 (DA3)~\cite{lin2025da3}, on the poses of a VIO filter, and address the remaining challenge of metric scale consistency. Although the poses are metric, the predicted depths retain image-dependent scale errors that accumulate across keyframe windows and cannot be revealed by their internal geometric consistency. In practice, these errors can reach a few percent: at a depth of three meters, a two-percent scale error amounts to six centimeters, already exceeding the five-centimeter tolerance used in our evaluation that uses F-Score at 5\,cm. Such errors are hidden when maps are evaluated after similarity alignment, but remain visible when placed using their own trajectories. We therefore aim to make the scale of every image metric and mutually consistent, and evaluate maps without fitting their scale to the reference as parts, but as a whole map including error from drifting.

We introduce \method, an online dense mapping system that addresses this problem at two levels. First, each keyframe window is anchored to metric 3D points triangulated using the filter's poses. These points are obtained from relative poses shared by the keyframes. Consequently, pose errors can shift them, making them mutually consistent while jointly biased. Each image can therefore retain a common scale error that cannot be detected from its own triangulated keypoints. The missing constraint comes from the map itself. We exploit bidirectional dense depth agreement between images from different windows observing the same surfaces to estimate and correct per-image scale errors. The agreement measures the difference between the scale errors of two images, while its sensitivity to pose errors depends on scene depth rather than the shorter triangulation baseline. From all such pairs, we solve one correction per image without labels, preserving the overall scale established by the metric anchors~(\cref{sec:anchor}). The correction requires no training or ground-truth depth and runs asynchronously without delaying mapping. The mapper never waits for the correction process and incorporates its latest available estimates, allowing the live map to remain scale-corrected. We further introduce context keyframes to improve the geometric predictions of the depth model.

\method is intended to be a universal solution that can be used on different everyday use devices, such as smartphones, to provide metric dense mapping. The mapper and depth model use a single configuration across different platforms. Each platform provides its rig file, including camera calibration, the camera-to-IMU transformation, and IMU noise parameters, as required by any VIO estimator. We evaluate \method with a single mapping configuration across five public camera--IMU setups and a new smartphone/GoPro dataset. \method achieves the best mean map F-score among the evaluated dense pipelines on every public dataset with a scan. On our smartphone/GoPro dataset, it achieves the best or second best mean score.

\paragraph{Contributions.}
\begin{itemize}[leftmargin=*,itemsep=1pt,topsep=2pt]
  \item \textbf{Scale-consistent mapping on a pose-conditioned depth model}: a metric anchor per window from keypoints on filtered depth; a label-free correction per image, measured by the dense multi-view two-way agreement of images that see one
  surface and solved online off the mapping path; and context keyframes for the depth model (\cref{sec:mapping}).
  \item \textbf{One configuration across platforms, scored at the map's own placement}: five public setups, against five other dense pipelines and two VIO sparse systems. Plus additional everyday-use devices (smartphones, GoPro) to ensure consistent robust dense metric mapping for general use cases.
  (\cref{sec:experiments}).
  \item \textbf{A smartphone rig dataset}: \phWalksWord{} campus walks by day and at night from two phones carried together, with a LiDAR phone's map as the reference (\cref{sec:exp_devices}), to be released.
\end{itemize}

\section{Related Work}\label{sec:related}

\paragraph{Feed-forward geometry and SLAM.}
DUSt3R~\cite{wang2024dust3r} and MASt3R~\cite{leroy2024mast3r} predict pairwise point maps, while VGGT~\cite{wang2025vggt} and $\pi^3$~\cite{wang2025pi3} jointly predict geometry and camera poses from multiple images. MapAnything~\cite{keetha2026mapanything} and DA3~\cite{lin2025da3} also support reconstruction conditioned on supplied camera intrinsics and poses. Building on feed-forward predictions, SLAM systems align geometry through $\Sim$ or $\mathrm{SL}(4)$ optimization~\cite{murai2025mast3rslam,maggio2025vggtslam,maggio2026vggtslam2,lee2026unisim,jeon2026aimslam}.
Joint pose and depth optimization is another central component of learned SLAM, as exemplified by DROID-SLAM~\cite{teed2021droid} and, more recently,
FoundationSLAM~\cite{wu2026foundationslam}, which incorporates a depth foundation model. We focus on inter-window depth-scale consistency in pose-conditioned mapping while retaining the metric trajectory supplied by a visual--inertial estimator.

\paragraph{Dense visual--inertial mapping with learned depth.}
Learned depth has been used within VIO for dense geometry estimation~\cite{zuo2021codevio} and
initialization~\cite{zhou2022learneddepthinit,zhang2026ffvioinit}.
Kimera~\cite{rosinol2020kimera} adds dense mapping to visual--inertial estimation, SimpleMapping~\cite{xin2023simplemapping} uses VIO poses and sparse points to guide learned multi-view stereo, and OKVIS2-X~\cite{jung2025uncertaintyvislam,boche2025okvis2x} couples uncertainty-aware
dense mapping with state estimation. MASt3R-Fusion~\cite{zhou2026mast3rfusion} integrates feed-forward visual alignment
with IMU and GNSS factors. VIDAR~\cite{salih2026vidar} anchors learned mapping to VIO, while ScaRF-SLAM~\cite{zhang2026scarfslam} conditions DA3 on tracking poses and optimizes frame and submap scales.
FoundationVIO~\cite{rhee2026foundationvio} incorporates relative poses predicted by a geometric foundation model into an inertial factor graph, jointly estimating
camera poses and a scale for each image batch.
Our method instead builds on the pose-conditioned mapping of DAVIO~\cite{davio2026},
retaining the filter's pose stream while adding metric depth anchoring and
inter-frame scale correction.

\paragraph{Metric anchoring and scale consistency.}
VI-Depth~\cite{wofk2023videpth} aligns monocular depth predictions to sparse
visual--inertial depth through scale and shift, while visual--inertial rescaling has also been used for autonomous aerial navigation~\cite{yang2025rescaler}.
VI3~\cite{lozano2026vi3} uses inertial cues to ground pretrained geometry,
and Scalix~\cite{barbaslaina2026scalix} incorporates learned scale uncertainty into monocular SLAM. Beyond metric anchoring, consistency across predictions is addressed through submap alignment~\cite{maggio2025vggtslam,maggio2026vggtslam2},
frame and submap scale optimization in ScaRF-SLAM~\cite{zhang2026scarfslam},
temporal consistency in video depth~\cite{luo2020consistent,kopf2021robust},
and bidirectional geometric constraints in FoundationSLAM~\cite{wu2026foundationslam}.
Following the sparse-to-dense alignment principle, we estimate a robust
window-level log-scale offset from triangulated metric points. We then use bidirectional dense log-depth comparisons across windows to correct residual per-image scale differences, holding camera poses fixed during this correction. The log-scale corrections sum to zero within each connected component, so their multiplicative factors have geometric mean one.

\section{System}\label{sec:base}
\begin{figure*}[t]
  \centering
  \includegraphics[width=0.95\textwidth]{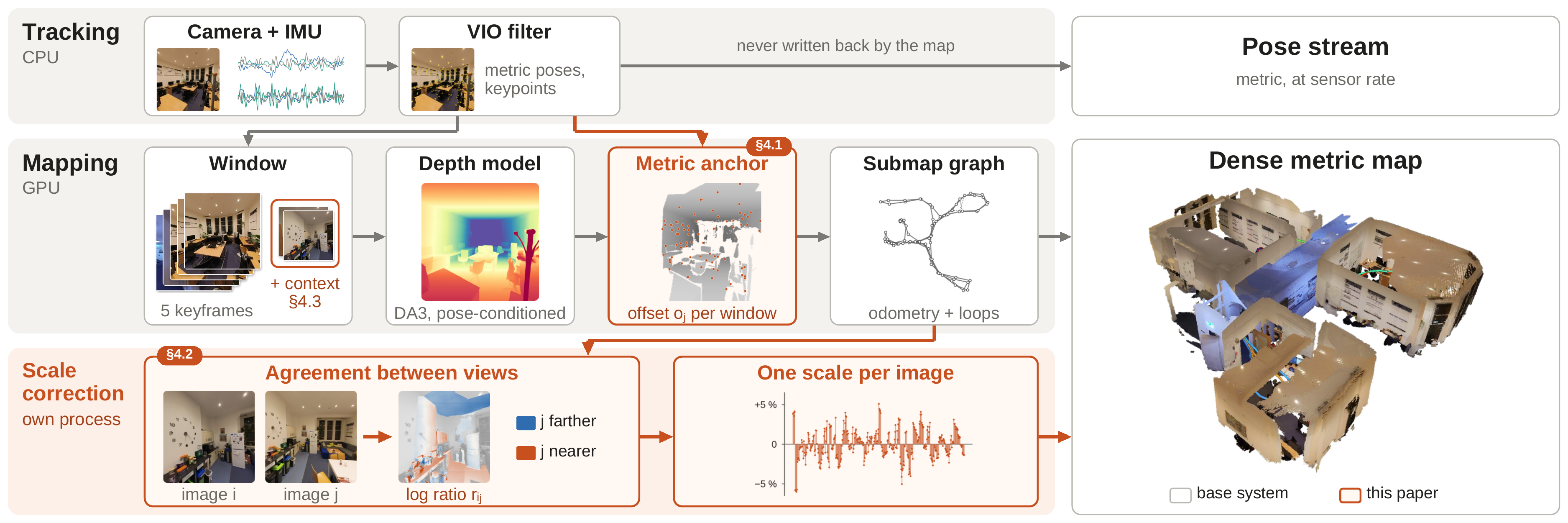}
  \caption{\textbf{Accord Architecture: three processes}. Gray: the base system~\cite{davio2026}; orange: \method (ours).}
  \label{fig:arch}
\end{figure*}
\paragraph{Overview.} $T_{AB}\in\SE$ maps coordinates in frame $B$ to frame $A$; $C_i$ is the camera at image $i$, $O$ is the odometry frame and $M$ is the map frame; 
The system runs as three processes (\cref{fig:arch}), and it is based on a dense monocular-inertial SLAM system with feed-forward initialization and pose-conditioning mapping called DAVIO~\cite{davio2026}. The filter-based VIO estimator~\cite{geneva2020openvins} is used for tracking visual-frames and fusing them with inertial data. This filter is used in~\cite{davio2026} to provide robust visual-inertial initialization and can provide robust and accurate relative transformation between covisible keyframes. The map then applies a rigid correction $T_{MO}$ to the filter relative poses. A single GPU worker runs DA3~\cite{lin2025da3} on \emph{windows} of $N{=}5$ keyframes, one keyframe every 0.5\,s, consecutive windows sharing two anchor keyframes. The scale correction of \cref{sec:revisit} then runs in the third process.

\paragraph{Pose-conditioned windows.} For window $j$, DA3 receives multi-view images, their intrinsics and their VIO filter's metric poses, and returns a depth map $d_{ji}$ and a confidence map for every image $i$, and its own estimate of the cameras. The similarity that best maps these cameras' centers onto the filter's ones gives the window's scale $\alpha_j$. The poses are already metric, so every scale is applied along the viewing rays only: a pixel $\mathbf u$ of
image $i$ is placed at:
\begin{equation}
  \mathbf p_M = T_{MS_j}\,T_{S_jC_i}\bigl(e^{\lambda_j - o_j - c_i}\,\alpha_j\,s_{ji}\,d_{ji}(\mathbf u)\,K^{-1}\tilde{\mathbf u}\bigr),
  \label{eq:placement}
\end{equation}
where $S_j$ is the window's center camera and $T_{MS_j}$ its pose in the map, $K$ the intrinsics, $\tilde{\mathbf u}$ the homogeneous pixel coordinates, $T_{S_jC_i}$ the filter's relative pose, $s_{ji}$ a relative scale per image solved within the window from matched keypoints with the poses fixed~\cite{zhang2026scarfslam}, and $\lambda_j$ the window's log-scale in the graph. The two terms this paper adds are the window's metric anchor $o_j$ (\cref{sec:anchor}) and the image's correction $c_i$ (\cref{sec:revisit}). Following scale correction, \method applies a depth filter through three steps:
\begin{itemize}
    \item \textbf{Image Quantile Confidence}: DA3 assigns a confidence score to each pixel, and the least confident fifth of every image is dropped.
    \item \textbf{Depth Edge Filtering}: within a $3 \times 3$ neighborhood around each pixel, the difference between the maximum and minimum depth must be below 15\% of the pixel’s depth. This helps to remove flying pixels hanging in mid-air between the object and the background.
    \item \textbf{Multi-View Consistency}: pixels are uplifted and projected into another image in the same window to check consistency.
\end{itemize}

\paragraph{Submap factor graph and loop closure.} Each window becomes a submap, a node $(T_{MS_j},\lambda_j)$ of a factor graph that optimizes its position, yaw, and $\lambda_j$; roll and pitch are kept fixed from the filter, which observes them via gravity~\cite{qin2018vinsmono}. Consecutive submaps are constrained by the filter’s relative pose using matched 3D points and the depth ratio of their shared images. For loop closure, nearby older submaps are tested as candidates: the relative pose between the two is estimated from the depth of one, verified against the depth of the other, and kept only if it agrees with the odometry within their joint uncertainty. A loop closure updates poses only where the odometry drift may exceed the loop’s own error; otherwise, only its depth ratio is used. The graph is re-optimized after each window. The live map fuses the latest submaps at their current poses on 4\,cm voxels in online mode, while the final map fuses all of them.

\paragraph{What a platform brings.} \method is designed as a universal dense mapping system used by and evaluated on various platforms. The mapper, the graph, and the depth model require no platform-specific parameter tuning. A new device requires rough estimates from its rig file: the camera's calibration (which can be estimated from DA3~\cite{lin2025da3}), camera-IMU extrinsics (refined online within the VIO filter~\cite{qin2018vinsmono}), sensor time offsets, and IMU noise parameters. Even when evaluated with inaccurate calibration data, \method still outperforms competing mapping systems under the same calibration conditions (Sec.~\ref{sec:exp_devices}).

\section{Scale-Consistent Mapping}\label{sec:mapping}

\subsection{Metric scale anchor}\label{sec:anchor}
A model conditioned on poses does not reliably adopt their scale, so each window must be aligned to sparse metric points following~\cite{wofk2023videpth,yang2025rescaler}. The keypoints of every image of window $j$ are matched to the center images of older covisible windows and triangulated to 3D map space. A keypoint that triangulates well becomes a target if it passes the three-step depth filter. For a target $k$ at pixel $\mathbf u_k$ of image $i$, with triangulated depth $z_k$, the log-depth residual and its weight are
\begin{equation}
\delta_k=\log\bigl(\alpha_j\,s_{ji}\,d_{ji}(\mathbf u_k)\bigr)-\log z_k,\qquad w_k=\frac{1}{v_k+\sigma_0^2},  \label{eq:residual}
\end{equation}
where $v_k=v_k^{\rm pix}+v_k^{\rm pose}$ is the variance of $\log z_k$: $v_k^{\rm pix}$ propagated from the noise of the matched pixels, $v_k^{\rm pose}$ from the error of the filter's relative position of the two cameras; $\sigma_0$ is a lower bound on each target's uncertainty. Given $\mathcal T_j$ as the window's targets, the anchor is the robust weighted mean of the residuals,
\begin{equation}
  o_j=\argmin_{o}\sum_{k\in\mathcal T_j}w_k\,\psi(\delta_k-o),
  \label{eq:anchor}
\end{equation}
with $\psi$ the Huber loss and $|o_j|$ bounded; it is removed from the log-depth of every image of the window (\cref{eq:placement}). One offset per window suffices because the matches inside the window already fix the relative scales $s_{ji}$ of its images (\cref{sec:base}). An error in the relative position of two cameras changes every depth triangulated between them by nearly the same factor, so it moves the targets together, not apart. That common shift passes into $o_j$, where no weighting can remove it (\cref{sec:revisit} corrects it); $v_k$ counts only the part that differs from target to target. A window none of whose images has enough targets keeps $o_j=0$. In the graph, a tight prior holds every $\lambda_j$ near zero, so the graph does not spread the anchored scales apart again.

\subsection{The agreement between views}
\label{sec:revisit}
\paragraph{What the anchor leaves.} The targets $\mathcal T_j$ of a window $j$ are triangulated from covisible keyframes. An error $\epsilon$ of a pose along a baseline $b$ moves all targets of the pair by the same relative amount $\epsilon/b$, i.e., they agree with each other and are wrong together. Every image therefore keeps a log-scale error $e_i$ of a few percent, common to its pixels and invisible in its residuals.

\paragraph{The measurement.} The map sees what the anchor misses. For two images $i,j$ of different windows observing the same surface, we unproject the pixels of $i$ with its depth $d_i$ (\cref{eq:placement} without $c_i$), transform them into $j$ via the map poses and compare the resulting depths against $d_j$ by taking the median over the pixels $\mathbf{u}$ of $i$ that project into $j$:
\begin{equation}
  \begin{aligned}
  r_{ij}&=\operatorname*{med}_{\mathbf u}\big[\log d_j(\pi_{ij}(\mathbf u))-\log z_{i\to j}(\mathbf u)\big],\\
  a_{ij}&=\tfrac12\,(r_{ij}-r_{ji}),
  \end{aligned}
  \label{eq:revisit}
\end{equation}
where $\pi_{ij}$ projects a pixel of $i$ into $j$ and $z_{i\to j}$ is its depth in $j$; pixels whose viewing rays differ by more than $30^\circ$ or whose depth ratio is far from one are left out. If the depths of $i$ and $j$ are overestimated by factors $e^{e_i}$ and $e^{e_j}$, carrying $i$ into $j$ gives $r_{ij}\approx e_j-e_i$, and the reverse gives the opposite,
$r_{ji}\approx e_i-e_j$. An occlusion gives the same sign both ways: the image that cannot see a point reads the surface
in front of it, so its measured depth is shorter. Taking half the difference preserves the scale error while canceling out occlusions: $a_{ij}\approx e_j-e_i$. This cancellation is exact only for two cameras equally far from the surface; if one camera is closer by a distance $D$ along the viewing direction at scene depth $z$, then $a_{ij}=(e_j-e_i)-\tfrac{D}{z}\,\tfrac{e_i+e_j}{2}$ to first order, and
the extra term is below a tenth of a percent at the errors we measure (\cref{sec:s_scale}). Moreover, this comparison is largely insensitive to pose errors: a position error $\epsilon$ changes a depth triangulated over a baseline $b$ by
$\epsilon/b$, but changes $a_{ij}$ only by $\epsilon/z$, and the scene is several meters away while the baselines are a few decimeters. This is why, for any camera, relative image scales are measured on the map rather than through odometry.

\paragraph{The correction.} Each pair of images $(i,j)$ from different windows that see enough of the same surface, in both directions, gives one measurement $c_j-c_i\approx a_{ij}$ of the difference between their log-scale corrections. Over all such pairs, we solve 
\begin{equation}
  \mathbf c=\argmin_{\mathbf c}\sum_{(i,j)}\psi\big(a_{ij}-(c_j-c_i)\big)+\sum_i c_i^2 ,
  \label{eq:revisit_correction}
\end{equation}
and multiply the depth of image $i$ by $e^{-c_i}$ (\cref{eq:placement}). The pair terms determine only the differences between corrections. The quadratic term, which weighs as much as one pair per image, determines their common level. Because each pair term depends only on $c_j-c_i$, summing the optimality conditions over a group of images linked by pairs gives $\sum_i c_i=0$. The group therefore keeps the mean scale that the anchor gave it, and the correction removes only the disagreement between the group's images.

\paragraph{Online, off the mapping path.} The scale correction runs in its own process (\cref{fig:arch}). The mapper sends this process each window's depth at half resolution and the current nodes of the graph, applies the newest corrections the process has published, and never waits for it. The process measures a pair again only when the placement of one image in the other's frame has moved, and back-projects an image only where the pair overlaps enough for the measurement to be reliable. The corrections trail the map by about one window, so the live map
shows the corrected geometry while the pose stream is unchanged.

\subsection{Context keyframes for the depth model}
\label{sec:context}
Each window's depth is predicted from its own keyframes alone, leaving nothing to tie it to previously observed geometry. \method adds to DA3's input up to two earlier keyframes, the \emph{context keyframes}, inspired by~\cite{jeon2026aimslam}. These keyframes are covisible with the window's center camera and located within a close radius. Because these keyframes are passed with their current poses, an older keyframe's pose deviates from the window by the odometry drift accumulated since then. We therefore select only keyframes that are 2 to 20\,s older than the window: old enough to provide a meaningful baseline, yet recent enough that the accumulated drift remains small (\cref{sec:s_config}) and can be corrected by the factor graph optimization.

\section{Experiments}\label{sec:experiments}
\begin{table}[t]
  \centering\small
  \caption{Platform, camera, sequences, total time and path}
  \label{tab:setups}
  \resizebox{\columnwidth}{!}{\setlength{\tabcolsep}{2.3pt}
\begin{tabular}{@{}lllcc@{}}
\toprule
Setup & Platform, scene & Camera & Seq. & Time, path \\
\midrule
EuRoC & drone, rooms & gray & 6 & 11\,min, 0.4\,km \\
TUM-VI & hand-held, lab & gray fisheye & 6 & 14\,min, 0.7\,km \\
ORI & hand-held, buildings & color fisheye & 5 & 11\,min, 0.5\,km \\
Hilti-Oxford & survey rig, heritage & gray fisheye & 5{+}5 & 26\,min, 1.0\,km \\
Oxford Spires & hand-held, outdoors & color fisheye & 3 & 28\,min, 1.7\,km \\
\midrule
Phone-rig (ours) & two smartphones, campus & color & 11+10 & 59\,min, \phKm\,km \\
GoPro (ours) & GoPro HERO13, mall & color & 1 & 12\,min, 0.6\,km \\
 
\bottomrule
\end{tabular}
}
\end{table}
\begin{table*}[t]
  \centering\small
  \caption{\textbf{The public benchmarks.} Means over each dataset's sequences; \textbf{bold}: best,
  \underline{underlined}: second best. Left: the dense map's F-score at 5\,cm. Right: the trajectory's error in cm. $\times$: every sequence failed (no output, or a trajectory more than 10\,m drifted off)}
  \label{tab:overview}
  \resizebox{\textwidth}{!}{\setlength{\tabcolsep}{5pt}
\begin{tabular}{@{}l cccc @{\hspace{14pt}} cccc@{}}
\toprule
& \multicolumn{4}{c}{Dense map: F-score at 5\,cm $\uparrow$} & \multicolumn{4}{c}{Trajectory error (cm) $\downarrow$} \\
\cmidrule(r{14pt}){2-5}\cmidrule{6-9}
& EuRoC & ORI & Hilti-Oxford & Oxford Spires & EuRoC & TUM-VI & Hilti-Oxford & Oxford Spires \\
System & \footnotesize\textit{drone, rooms} & \footnotesize\textit{hand-held, buildings} & \footnotesize\textit{survey rig} &
\footnotesize\textit{outdoors} & \footnotesize\textit{drone, rooms} & \footnotesize\textit{hand-held} &
\footnotesize\textit{survey rig} & \footnotesize\textit{outdoors} \\
\midrule
ORB-SLAM3~\cite{campos2021orbslam3} & -- & -- & -- & -- & \textbf{3.2} & \underline{2.0} & 68.2 & -- \\
OKVIS2-X~\cite{boche2025okvis2x} & -- & -- & -- & -- & \underline{6.0} & \textbf{1.4} & \textbf{17.1} & -- \\
MASt3R-Fusion~\cite{zhou2026mast3rfusion} & 0.41 & 0.53 & \underline{0.13} & $\times$ & 7.0 & 3.0 & 83.7 & -- \\
VGGT-SLAM~2.0~\cite{maggio2026vggtslam2}$^\dagger$ & \underline{0.49} & 0.18 & 0.04 & 0.05 & 7.4 & 25.0 & 438.3 & 382.5 \\
ScaRF-SLAM's mapper~\cite{zhang2026scarfslam}$^\ddagger$ & 0.20 & \underline{0.74} & 0.09 & 0.05 & -- & -- & -- & -- \\
FoundationVIO~\cite{rhee2026foundationvio} & $\times$ & 0.56 & 0.06 & \underline{0.12} & -- & -- & 216.1 & 105.0 \\
DAVIO~\cite{davio2026} & 0.38 & 0.50 & -- & $\times$ & 7.7 & 5.8 & -- & \textbf{69.9} \\
\midrule
\method & \textbf{0.57} & \textbf{0.78} & \textbf{0.16} & \textbf{0.20} & 6.8 & 4.5 & \underline{37.0} & \underline{73.0} \\
 
\bottomrule
\end{tabular}
}
\end{table*}
\begin{figure*}[t]
  \centering
  \includegraphics[width=\textwidth]{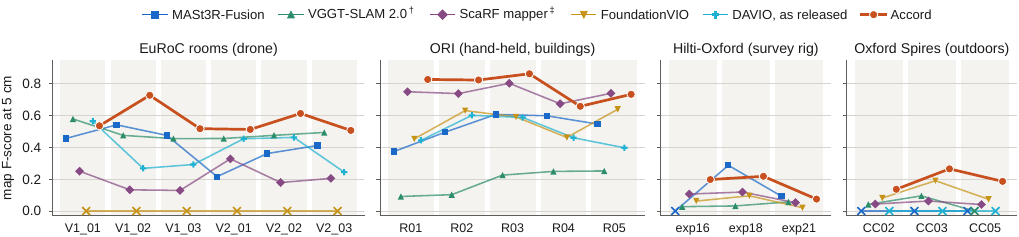}
  \caption{\textbf{Every scanned sequence.} The map's F-score at 5\,cm per sequence, placed by each system's own trajectory; the means are \cref{tab:overview}'s. A $\times$ at 0: no map, or the map of a trajectory that
  failed (more than 10\,m off). $^\dagger$, $^\ddagger$ as in \cref{tab:overview}.}
  \label{fig:spectrum}
\end{figure*}

We evaluate on five public benchmarks  (\cref{tab:setups}): a drone flying indoors
(EuRoC~\cite{burri2016euroc}), hand-held fisheye cameras in buildings (ORI~\cite{zhang2026scarfslam}) and in a motion-capture room (TUM-VI~\cite{schubert2018tumvi}), a survey rig in a historic building and on a construction site
(Hilti-Oxford~\cite{zhang2023hiltioxford}, where maps are scored in the indoor building sequences with available ground-truth scans,
\cref{sec:s_hilti}), and a rig carried through college quads outdoors, beyond the depth model's range (Oxford Spires~\cite{tao2025spires}): \setSeqN{} sequences over \setMinutes{} minutes and \setKm\,km.

\paragraph{Everyday devices.} The public benchmarks rely on lab and survey rigs. To test the system on consumer hardware that people carry daily, we add two custom setups: two smartphones carried across a university campus, with reference geometry captured via iPhone LiDAR scanning, and a GoPro carried through a shopping mall. Neither were calibrated on a target: our goal is an open-source system that performs online metric dense mapping from any device equipped with a single camera and IMU, given only a rough initial calibration.

\paragraph{One configuration.} All hyperparameters were selected on three development sequences from EuRoC and ORI (V1\_01, V1\_02, R01) and kept fixed across every other sequence and platform. A platform provides only its rig file, from the dataset, from a Kalibr~\cite{furgale2013kalibr} calibration or, for our own devices, from nominal values. On Hilti-Oxford the IMU's noise was measured from the release's recording at rest (the densities at their Allan values, the bias walks at three times their bounds), as required for its evaluation at surveyed fixed control points~\cite{zhang2023hiltioxford}.

\paragraph{Runs and metrics.} The system runs live on one RTX~4090 with DA3-Giant. Each of \method's reported results is the mean of a sequence's first three runs. Reconstructed maps are evaluated using the F-score at 5\,cm against the reference scan, on 4\,cm voxels, where accuracy is measured around the reference camera viewpoints and completeness is evaluated over the observed reference surface. Each map is
\emph{placed by its own trajectory} using the rigid alignment of the estimated trajectory to the reference without fitting to the surface, so a wrong scale or a drifting trajectory is directly penalized in the map's score. Trajectories are evaluated via ATE after a rigid alignment, without scale fitting; on Hilti-Oxford, by the error at its surveyed fixed points.

\paragraph{Other pipelines.} We compare against five baselines: MASt3R-Fusion~\cite{zhou2026mast3rfusion}, a feed-forward visual--inertial system evaluated with the dense map of its global optimization; VGGT-SLAM~2.0~\cite{maggio2026vggtslam2}, a vision-only method rescaled using the reference scale prior to the same rigid alignment; the mapper of ScaRF-SLAM~\cite{zhang2026scarfslam}, run with its own depth model, on \method's poses. FoundationVIO~\cite{rhee2026foundationvio}, a concurrent method evaluated with its dense map; DAVIO~\cite{davio2026}, our
base system paired with the same depth model. All baselines are run using their released configurations on the same images.
\subsection{Public benchmarks}
\label{sec:exp_map}\label{sec:exp_further}
\begin{figure*}[t]
  \centering
  \includegraphics[width=0.85\textwidth]{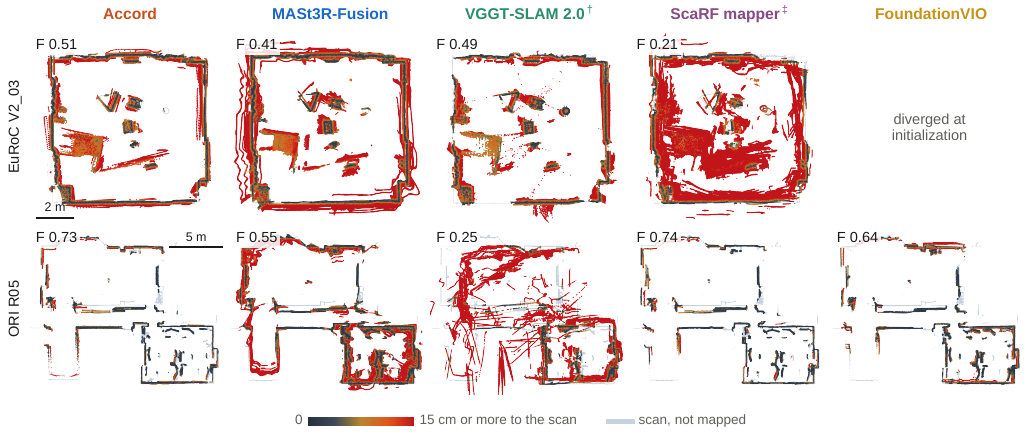}
  \caption{\textbf{Maps against the laser scan.} A top-view horizontal slice of each pipeline's map (light blue: scanned but not mapped). Every map point is colored by its distance to the scan. F is the sequence's F-score at 5\,cm}
  \label{fig:maps}
\end{figure*}
\begin{figure}[t]
  \centering
  \includegraphics[width=\columnwidth]{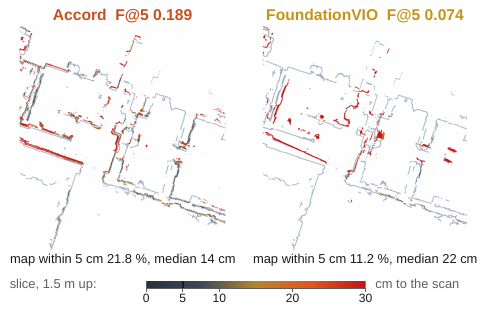}
  \caption{\textbf{\method{} against FoundationVIO outdoors} (Oxford Spires CC05).}
  \label{fig:spires}
\end{figure}
\paragraph{Maps.} With a single configuration, \method's map has the best mean across all four datasets with ground-truth scans
(\cref{tab:overview}). Notably, the runner-up is a different pipeline in each case: VGGT-SLAM~2.0 on EuRoC, ScaRF-SLAM's mapper on ORI, MASt3R-Fusion on Hilti-Oxford, and FoundationVIO on Oxford Spires. Each baseline excels in a specific type of scene, yet none ranks among the top two on a second dataset; what sets \method apart is its consistent generalization across platforms. It also holds per sequence (\cref{fig:spectrum}): \method's map has the highest score on \specOursLead{} of the \specSeqN{}
scanned sequences and one of the two highest on \specOursTopTwo. Every pipeline scores far lower on Hilti-Oxford and Oxford Spires than in the rooms and on ORI; confirming that large-scale heritage and outdoor environments remain challenging for every method, \method included.

\paragraph{Where the maps differ.} \Cref{fig:maps} shows what the scores measure. In the room, \method's walls align closely with the reference scan, whereas MASt3R-Fusion produces duplicated walls and ScaRF-SLAM's mapper scatters noisy surfaces across the room. On ORI ScaRF-SLAM's mapper, supplied with our poses, matches our reconstruction quality; FoundationVIO leaves part of the building unmapped, and VGGT-SLAM~2.0's map fractures along the trajectory. Outdoors (\cref{fig:spires}), FoundationVIO maps the ground along its path, where its depth is kept, while \method{} reaches the facades across the quad.

\paragraph{Concurrent work.} ScaRF-SLAM~\cite{zhang2026scarfslam} and FoundationVIO~\cite{rhee2026foundationvio}
evaluate maps under a local alignment protocol: the path is cut into 10\,m pieces and each piece is registered to the scan before it is scored, which masks accumulated drift. In contrast, our protocol aligns the entire map once using the system's own trajectory, ensuring that drift is directly reflected as map error. Under ours, FoundationVIO scores \fvioOriMap{} on ORI, behind \method{} on all \fvioOriAheadOurs{} sequences, and \spFvioMap{} against our \spOursMap{} on Oxford Spires, behind on all \spAheadMap; in every EuRoC room it diverges at its initialization, which fixes the scale before the drone has moved. Under theirs, with the recall rule its paper states, FoundationVIO's paper reports \fvioScarfPaper\,\% on ORI, our run of its release reaches \fvioScarfRunCut\,\% and \method{} \oursScarfCut\,\% (\cref{sec:s_fvio}).

\subsection{Everyday devices}\label{sec:exp_devices}
\begin{figure*}[t]
  \centering
  \includegraphics[width=0.9\textwidth]{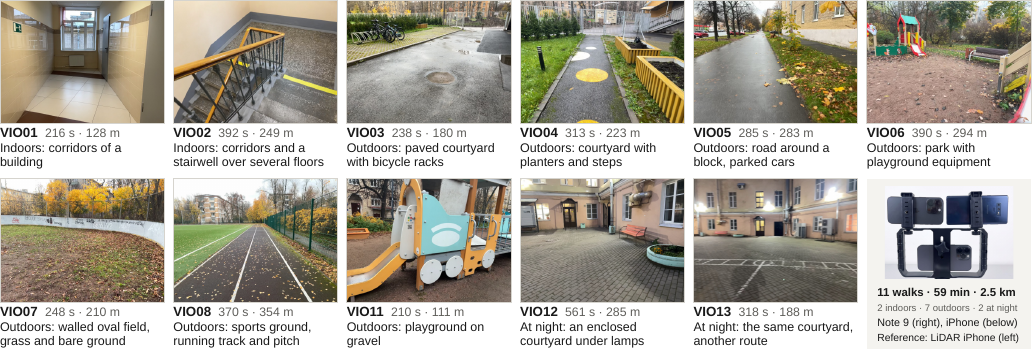}
  \caption{\textbf{The phone rig dataset.} The \phWalksN{} campus walks span indoor and outdoor environments, including two walks at night. Bottom
  right: the rig. The Note~9 (right) and the LiDAR iPhone that runs RTAB-Map~\cite{labbe2019rtabmap} (left), whose result is the reference. The phones are carried by hand on this frame and are not calibrated (\cref{sec:s_phone_rig}).}
  \label{fig:phone-dataset}\label{fig:phones}
\end{figure*}
\begin{table}[t]
  \centering\small
  \caption{\textbf{The phone benchmark.} Means over each phone's recordings: map F-score at 5 and 10\,cm, ATE(m).}
  \label{tab:devices}
  \resizebox{\columnwidth}{!}{\setlength{\tabcolsep}{3.2pt}
\begin{tabular}{@{}l ccc @{\hspace{8pt}} ccc@{}}
\toprule
& \multicolumn{3}{c}{Note~9, \phAndroidN{} walks} & \multicolumn{3}{c}{iPhone, \phIphoneN{} walks} \\
\cmidrule(r{8pt}){2-4}\cmidrule{5-7}
System & F@5 $\uparrow$ & F@10 $\uparrow$ & ATE (m) $\downarrow$ & F@5 $\uparrow$ & F@10 $\uparrow$ & ATE (m) $\downarrow$ \\
\midrule
ORB-SLAM3~\cite{campos2021orbslam3} & -- & -- & \textbf{0.71} & -- & -- & \underline{0.88} \\
OKVIS2-X~\cite{boche2025okvis2x} & -- & -- & -- & -- & -- & \textbf{0.34} \\
MASt3R-Fusion~\cite{zhou2026mast3rfusion} & 0.04 & 0.08 & 1.88 & $\times$ & $\times$ & -- \\
VGGT-SLAM~2.0~\cite{maggio2026vggtslam2}$^\dagger$ & 0.07 & 0.14 & 4.76 & 0.09 & 0.16 & 2.72 \\
ScaRF-SLAM's mapper~\cite{zhang2026scarfslam}$^\ddagger$ & 0.14 & 0.26 & -- & 0.10 & 0.18& -- \\
FoundationVIO~\cite{rhee2026foundationvio} & \underline{0.15} & \underline{0.27} & 2.10 & \textbf{0.13} & \textbf{0.22} & 1.57 \\
DAVIO~\cite{davio2026} & 0.12 & 0.23 & \underline{0.94} & 0.11 & 0.19 & 1.05 \\
\midrule
\method & \textbf{0.21} & \textbf{0.36} & \underline{0.94} & \underline{0.12} & \underline{0.21} & 1.06 \\
 
\bottomrule
\end{tabular}
}
\end{table}
\begin{figure}[t]
  \centering
  \includegraphics[width=\columnwidth]{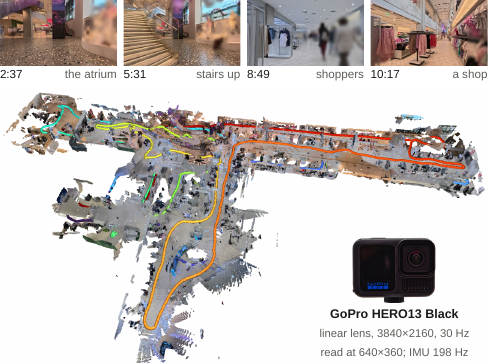}
  \caption{\textbf{A GoPro through a shopping mall.}}
  \label{fig:gopro}
\end{figure}
\paragraph{The phone rig.} Our dataset comprises \phWalksN{} campus walks: \phIndoorN{} indoors, \phOutdoorN{} outdoors by day and \phNightN{} at night (\phMinutes{} minutes, \phKm\,km in total; \cref{fig:phone-dataset}). A Samsung Note~9 and an iPhone were mounted together on a handheld rig alongside a LiDAR-equipped iPhone running RTAB-Map~\cite{labbe2019rtabmap} (\cref{fig:phone-dataset}, bottom right), whose loop-closed graph and map are the reference. The phones are uncalibrated. Maps and trajectories are scored as on the public benchmarks, with map points farther than 3\,m from the reference trajectory excluded.

\paragraph{Results on the phones.} All pipelines score far lower here than on the public benchmarks
(\cref{tab:devices}), which shows their sensitivity to a coarse calibration: the intrinsics come from the phones' metadata and ARKit's estimates, and the IMU's noise is assumed (\cref{sec:s_phone_rig}). Our map has the best mean on the Note~9, the highest score on \phAndroidLead{} of \phAndroidN{} recordings; on the iPhone, FoundationVIO achieves a slightly higher mean, with \method following closely behind.

\paragraph{The GoPro.} A GoPro HERO13 Black was carried by hand for \gpMinutes{} minutes and \gpMetres\,m over two levels of a busy shopping mall (\cref{tab:setups}), recording 4K video at \gpFrameHz\,Hz through its linear lens and
its own IMU at \gpImuHz\,Hz; \cref{fig:gopro}. The system reads the video at $640{\times}360$. No calibration target was used: the intrinsics follow from the lens's nominal field of view, the camera's pose on the IMU is taken as the identity and the IMU's noise as a phone's. Dynamic pedestrians are masked out from both tracking and mapping using a YOLO person detector. This experiment provides a qualitative evaluation, as no ground-truth point cloud scan is available.

\subsection{Ablation}\label{sec:exp_ablation}
\begin{table}[t]
  \centering\small
  \caption{\textbf{Ablation}: the parts of our mapper added one at a time, with DA3-Giant showing Mean F-score at 5\,cm of the map.}
  \label{tab:ablation}
  \resizebox{\columnwidth}{!}{\begin{tabular}{@{}lcc@{}}
\toprule
 & EuRoC rooms & Hilti-Oxford \\
\midrule
DAVIO~\cite{davio2026}& 0.381 & -- \\
our mapper, no anchor or context & 0.488 & 0.120 \\
\quad + the anchor & 0.547 & 0.147 \\
\quad + context keyframes (\method) & 0.570 & 0.164 \\
 
\bottomrule
\end{tabular}
}
\end{table}
\Cref{tab:ablation} starts from our mapper without the anchor and the context keyframes and adds them one at a time.

\paragraph{The anchor} provides the largest improvement increasing the mean corrected F-score on EuRoC Rooms from \cumNoneRoomsCorr{} to \cumAnchorRoomsCorr{},
and on Hilti-Oxford from \cumNoneHiltiCorr{} to \cumAnchorHiltiCorr{}.

\paragraph{Context keyframes} raise the corrected map score on \cumSystemAheadRooms{} of \resRoomsN{} on EuRoC Rooms
(\cumAnchorRoomsCorr{} to \cumSystemRoomsCorr) and on \cumSystemAheadHilti{} of 3 Hilti-Oxford sequences
(\cumAnchorHiltiCorr{} to \cumSystemHiltiCorr).

\subsection{The depth model}\label{sec:exp_model}
\paragraph{Depth model.} The mapper is agnostic to the backbone size of the depth model, and scaling up the model yields higher reconstruction accuracy across both datasets (\cref{tab:model}).
\begin{table}
  \centering\small
  \caption{\textbf{Three depth models.} F-score at 5\,cm.}
  \label{tab:model}
  \begin{tabular}{@{}lccc@{}}
\toprule
Depth model & EuRoC Rooms, F & ORI, F & GPU (GB) \\
\midrule
DA3-Giant & 0.570 & 0.782 & 8.9 \\
DA3-Large & 0.521 & 0.698 & 5.1 \\
DA3-Base & 0.499 & 0.677 & 2.5 \\
 
\bottomrule
\end{tabular}

\end{table}

\paragraph{Latency and compute.}
Table~\ref{tab:speed_memory} shows the latency and compute of \method against other systems in terms of compute, memory and speed.
\begin{table}[t]
  \centering\small
  \caption{\textbf{Speed and memory} on EuRoC, one RTX~4090.}
  \label{tab:speed_memory}
  \setlength{\tabcolsep}{4pt}
  \begin{tabular}{@{}lccc@{}}
    \toprule
    System & Speed (FPS) & GPU (GB) & RAM (GB) \\
    \midrule
    \method             & 18 & 11.1 & 2.4 \\
    FoundationVIO       & 15 & 13.2 & 7.1 \\
    MASt3R-Fusion       & 12 & 4.9  & 6.8 \\
    VGGT-SLAM~2.0       & 23 & 6.8  & 9.9 \\
    ScaRF-SLAM's mapper & 15 & 9.4  & 5.6 \\
    \bottomrule
  \end{tabular}
\end{table}

\section{Conclusion}\label{sec:conclusion}
A camera and an IMU are present on nearly every mobile device. Using a single fixed configuration, \method{} reconstructs dense metric maps online—anchored in scale to the filter's sparse keypoints and made globally consistent through inter-view depth agreement across covisible surfaces. From a drone flying indoors to a multi-camera rig carried across college quads, \method{} achieves the highest mean reconstruction score among all evaluated dense pipelines on every benchmark with ground-truth scans. Even on handheld smartphones walking across a campus with only coarse initial calibration, it produces accurate dense metric maps, establishing \method{} as a versatile, general-purpose system for online dense mapping.

{
  \small
  \bibliographystyle{ieeenat_fullname}
  \bibliography{ref}
}
\newpage
\FloatBarrier
\appendix
\section*{\centering\Large Appendix}
\raggedbottom
\renewcommand{\dblfloatpagefraction}{0.6}

\noindent This appendix gives the details the main text has no room for, in four parts. \emph{The method:}
\cref{sec:s_correction} explains the scale correction step by step. \emph{The setup:} \cref{sec:s_config} gives
the configuration and the evaluation protocol. \emph{The results of every sequence:} EuRoC and ORI
(\cref{sec:s_results}), Hilti-Oxford and TUM-VI (\cref{sec:s_hilti}), Oxford Spires (\cref{sec:s_spires}),
FoundationVIO against its own paper (\cref{sec:s_fvio}), and the phone rig dataset and the GoPro walk
(\cref{sec:s_devices}). \emph{Further tests:} the depth model (\cref{sec:s_ablation}), compute
(\cref{sec:s_compute}) and the limits of the system (\cref{sec:s_limits}).

The main text points here for five things: for the size of the term that cameras at unequal distances
add to the measurement (\cref{sec:s_measure}), for the age and the drift of the context keyframes
(\cref{sec:s_config}, \emph{Context keyframes}), for the Hilti-Oxford sequences whose maps are scored
(\cref{sec:s_hilti}), for FoundationVIO under the protocol of its paper (\cref{sec:s_fvio}), and for the
calibration of the phones (\cref{sec:s_devices}, \emph{The rig files}).

\section{The scale correction in  detail}\label{sec:s_correction}\label{sec:s_scale}\label{sec:s_online}
\Cref{sec:revisit} states the correction in three equations. This section explains each step, gives the values the implementation uses, shows the effect on one map (\cref{fig:s_floor}), and describes how the correction runs online.

\subsection{What is left to correct}
\begin{figure}[t]
  \centering
  \includegraphics[width=\columnwidth]{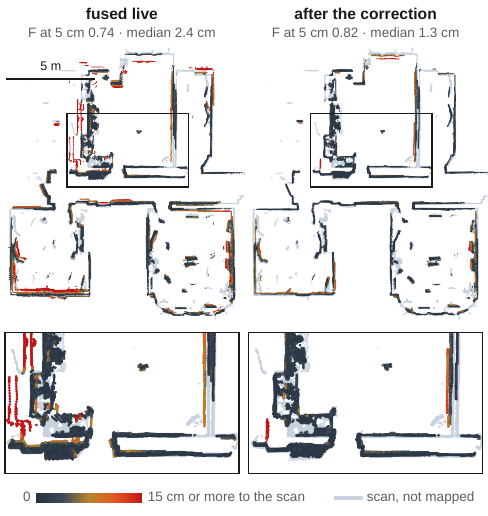}
  \caption{\textbf{Before and after the correction} (ORI R02, a live run): a horizontal slab of one floor from above, placed by the system's own trajectory. Every map point is colored by its distance to the scan, drawn under the map. Below, the framed part enlarged. As built (left), images that disagree in scale by a few percent smear a wall into layers. After the correction they agree.}
  \label{fig:s_floor}
\end{figure}

After the anchor (\cref{sec:anchor}), the depth of image $i$ is still wrong by one factor $e^{e_i}$, common to all its pixels: $e_i$ is the image's \emph{log-scale error}. The anchor cannot see $e_i$. Its targets are triangulated from the filter's poses, and an error of those poses moves all targets of an image by the same relative amount, so the targets agree with each other and with the depth they were used to scale. 
Two images that see one wall with different errors draw it twice (\cref{fig:s_floor}).

\subsection{The measurement, step by step}\label{sec:s_measure}
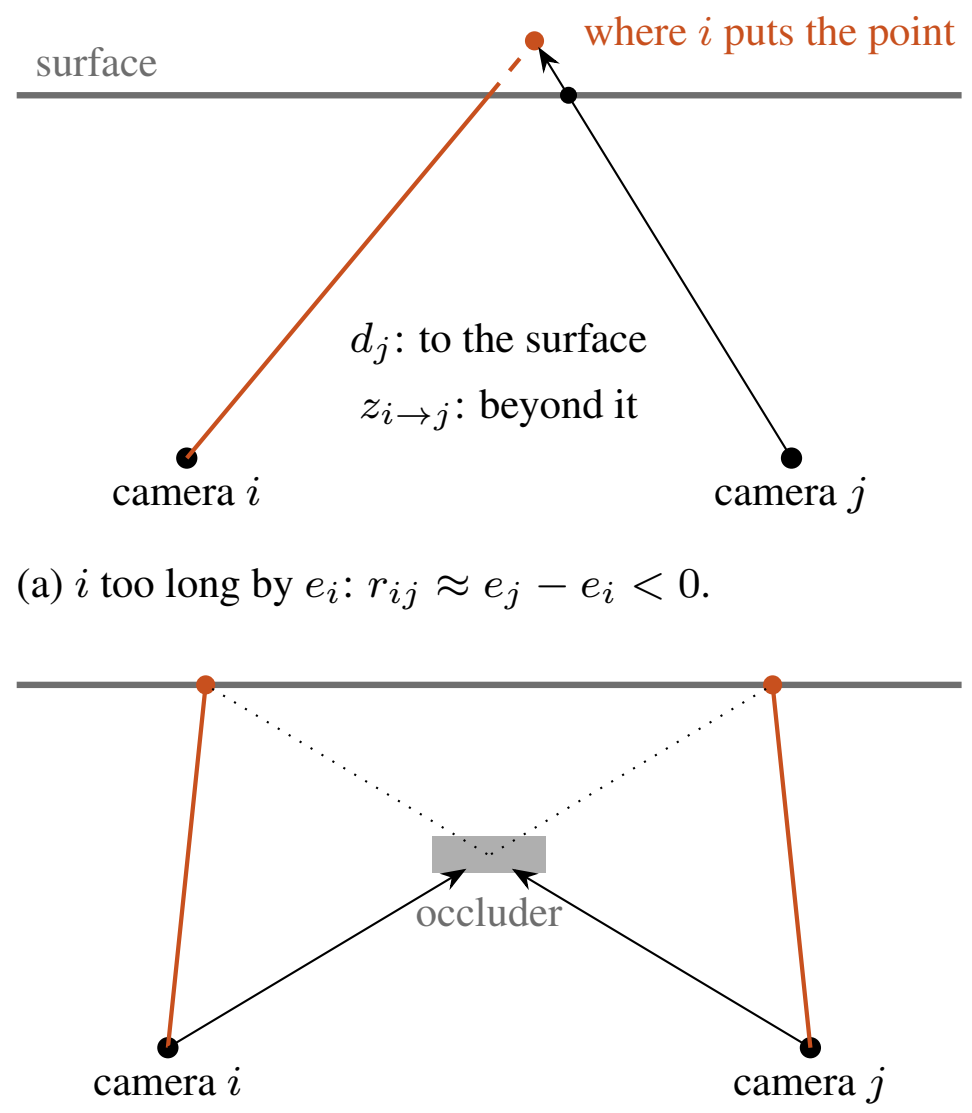
\begin{figure}[t]
  \centering
  \begin{tikzpicture}[x=1.25cm,y=0.8cm,font=\footnotesize,>=Stealth]
    \begin{scope}
      \draw[line width=1.2pt,basegray] (0,3)--(5,3);
      \node[basegray,anchor=south west] at (0,3) {surface};
      \fill (0.9,0) circle (2pt) node[below] {camera $i$};
      \fill (4.1,0) circle (2pt) node[below] {camera $j$};
      \draw[teach,thick] (0.9,0)--(2.5,3);
      \draw[teach,thick,dashed] (2.5,3)--(2.74,3.45);
      \fill[teach] (2.74,3.45) circle (1.8pt);
      \node[teach,anchor=west] at (2.9,3.5) {where $i$ puts the point};
      \draw[->,thin] (4.1,0)--(2.76,3.40);
      \fill (2.92,3) circle (1.6pt);
      \node at (2.55,0.95) {$d_j$: to the surface};
      \node at (2.55,0.4) {$z_{i\to j}$: beyond it};
      \node[anchor=north west] at (-0.1,-0.75) {(a) $i$ too long by $e_i$: $r_{ij}\approx e_j-e_i<0$.};
    \end{scope}
    \begin{scope}[yshift=-3.9cm]
      \draw[line width=1.2pt,basegray] (0,3)--(5,3);
      \fill[basegray!55] (2.2,1.45) rectangle (2.8,1.75);
      \node[basegray,anchor=north] at (2.5,1.4) {occluder};
      \fill (0.8,0) circle (2pt) node[below] {camera $i$};
      \fill (4.2,0) circle (2pt) node[below] {camera $j$};
      \draw[teach,thick] (0.8,0)--(1.0,3);
      \fill[teach] (1.0,3) circle (1.8pt);
      \draw[->,thin] (4.2,0)--(2.62,1.48);
      \draw[thin,dotted] (2.49,1.6)--(1.0,3);
      \draw[teach,thick] (4.2,0)--(4.0,3);
      \fill[teach] (4.0,3) circle (1.8pt);
      \draw[->,thin] (0.8,0)--(2.38,1.48);
      \draw[thin,dotted] (2.51,1.6)--(4.0,3);
      \node[anchor=north west] at (-0.1,-0.75) {(b) each reads the occluder: $r_{ij}<0$ and $r_{ji}<0$.};
    \end{scope}
  \end{tikzpicture}
  \caption{\textbf{The two-way measurement.} (a)~A scale error: image $i$ puts a surface point too far; seen from $j$, the point lies behind the surface that $j$'s own depth $d_j$ reads. Carried the other way the sign flips.  (b)~An occlusion: a point one camera sees is hidden from the other, which reads the nearer occluder. The sign is the same both ways, so half the difference of the two directions keeps the scale error and removes what the occlusions share.}
  \label{fig:s_twoway}
\end{figure}
Take two images $i$ and $j$ of different windows that see the same surface (\cref{fig:s_twoway}a).

\emph{One direction.} Each pixel $\mathbf u$ of $i$ is lifted to 3D with $i$'s depth and carried into $j$ with the map's poses. There it has a depth $z_{i\to j}(\mathbf u)$ and falls on a pixel of $j$, where $j$'s own depth map reads $d_j$. If $i$'s depth is too long, the point lies behind the true surface, so $z_{i\to j}$ is too long by about $e_i$ in log. $j$'s own reading of that surface is too long by $e_j$. The log ratio of the two, taken as a median over the pixels of $i$ that land in $j$, is therefore \[ r_{ij}\approx e_j-e_i . \]

\emph{The other direction.} Carrying $j$ into $i$ gives $r_{ji}\approx e_i-e_j$, the same number with the opposite sign.

\emph{Why both.} Where a point that $i$ sees is hidden from $j$, the pixel of $j$ shows the surface in front, and $j$'s depth there is shorter than $z_{i\to j}$ (\cref{fig:s_twoway}b). This pulls $r_{ij}$ down, and for the same reason it pulls $r_{ji}$ down: an occlusion has the same sign in both directions, a scale error opposite signs. Writing the occlusions' pull as $b_{ij},b_{ji}\le 0$,
\[ a_{ij}=\tfrac12(r_{ij}-r_{ji})\approx(e_j-e_i)+\tfrac12(b_{ij}-b_{ji}), \]
so the part of the pull that the two directions share cancels, and the scale difference stays. A median and a gate remove most of the pull before that: a pixel whose log ratio is beyond 0.5 is treated as an occlusion and left out.

\emph{Cameras at unequal distances.} The cancellation above is exact for two cameras equally far from the surface. If $j$ is nearer by $D$ along the view at scene depth $z$, the displacement $e_i z$ of a point of $i$ is a larger share of $j$'s shorter depth, $r_{ij}\approx e_j-e_i(1+D/z)$, and the reverse gives $r_{ji}\approx e_i-e_j(1-D/z)$, so 
\[ a_{ij}=(e_j-e_i)-\frac{D}{z}\,\frac{e_i+e_j}{2}. \]
The extra term scales with the \emph{mean} error of the two images, not with their spread, and the anchor holds this mean near the scale error of the filter. That error is small: over the anchor's baselines, the filter's displacements differ from the reference's by $\stRoomsFilter$\,\% on the EuRoC rooms and by $\stOriFilter$\,\% on ORI. For a camera that advanced by a tenth of the scene depth, the extra term is a tenth of these values: about a tenth of a percent on ORI and less in the rooms. The corrections themselves are more than ten times larger, with a median size of \corrAbsMedianRooms\,\% on the rooms and \corrAbsMedianOri\,\% on ORI, so the term is left out.

\emph{Pose errors.} A position error $\epsilon$ between the two cameras changes $a_{ij}$ by $\epsilon/z$, with $z$ the scene depth of several meters. The same error changes a depth triangulated over a baseline $b$ of a few decimeters by $\epsilon/b$, roughly ten times more. This is why the relative scales of images are measured on the map's surfaces and the absolute scale is left to the anchor.

\emph{Which pairs count.} The source pixels are every sixth row and column of an image, restricted to the depth the three-step filter kept. A pixel takes part if it lands inside the other image at more than 0.15\,m, the two viewing rays of the point are within $30^\circ$, and its log ratio is within 0.5. A pair of images counts only if they belong to different windows and at least 500 pixels take part in \emph{each} direction (\cref{tab:s_config}). Images of one window are not paired: their relative scales $s_{ji}$ are already solved inside the window.

\subsection{The solve, and what it keeps}
Every counted pair gives one equation $c_j-c_i\approx a_{ij}$ for the corrections $\mathbf c$
(\cref{eq:revisit_correction}). The Huber loss is minimized by ten rounds of reweighted least
squares: with weights $w_{ij}$ (one at first), each round solves the linear system
\[ (L+I)\,\mathbf c=\mathbf b,\qquad b_k=\sum_{i}w_{ik}a_{ik}-\sum_{j}w_{kj}a_{kj}, \]
where $L$ is the weighted Laplacian of the graph whose nodes are the images and whose edges are the pairs, and $I$ is the quadratic term. After each round, a pair whose residual exceeds 1.5 robust standard deviations of all residuals is weighted down in proportion.

\emph{The mean is kept.} Each row of $L$ sums to zero, and each pair adds $+w_{ij}a_{ij}$ to one entry of $\mathbf b$ and $-w_{ij}a_{ij}$ to another. Summing the system over a group of images linked by pairs therefore gives $\sum_i c_i=0$: the corrections of a group cancel on average. The group keeps the mean scale the anchor gave it, and only the disagreement between its images is removed. The correction cannot drift the map's scale.

\emph{Few pairs, small correction.} The quadratic term weighs as much as one pair. An image with $k$ pairs moves $k/(k+1)$ of the way to where its pairs put it: half the way with one pair, nearly all of it with ten. An image with no pair keeps $c_i=0$.

\subsection{Online mapping}
The correction runs in a process of its own, beside the mapper (\cref{fig:arch}). After each window the mapper sends this process the window's filtered depth at half resolution and the current nodes of the graph, and takes the newest corrections the process has published. It never waits for them. Two rules keep the process fast without changing its result.
\begin{itemize}
    \item A pair of windows is measured again only when the placement of one in the other's frame has moved by more than 2\,cm (or by a rotation or a scale change worth as much at 3\,m). A shift of the whole map is no move.
    \item An image is carried back into another only if the first direction already found the 500 shared pixels the pair needs.
\end{itemize}
The corrections trail the map by about one window, and the last ones arrive \pobDrainLo{} to \pobDrainHi\,s after a sequence ends.

\section{Configuration and protocol}\label{sec:s_config}\label{sec:s_protocol}
\paragraph{One configuration.} \Cref{tab:s_config} lists the settings of the system. They were chosen on the three development sequences (V1\_01, V1\_02 and R01) and are the same on every sequence and platform. Per platform, only the rig file changes (the camera's calibration, its pose on the IMU and the IMU's noise) and, on Hilti-Oxford alone, the filter is also updated while the rig stands still
(\cref{sec:s_hilti}). The filter is OpenVINS~\cite{geneva2020openvins} with its own initialization. It refines the camera--IMU extrinsics online. The depth model is given frames undistorted to a pinhole view (for a fisheye, a square of $90^\circ$ and 512 pixels) and the gray image the filter tracks. Keypoints are XFeat~\cite{potje2024xfeat}, matched by its LighterGlue.
\begin{table}[t]
  \centering\footnotesize\renewcommand{\arraystretch}{0.92}%
  \caption{\textbf{The settings of the system}, the same on every platform.}
  \label{tab:s_config}
  \begin{tabular}{@{}lr@{}}
\toprule
Setting & Value \\
\midrule
\multicolumn{2}{l}{\emph{Windows}} \\
\quad keyframes per window & 5 \\
\quad keyframes two windows share & 2 \\
\quad seconds between keyframes & 0.5 \\
\quad seconds a window may wait for the worker & 2 \\
\multicolumn{2}{l}{\emph{Context keyframes}} \\
\quad at most & 2 \\
\quad youngest (s before the window) & 2 \\
\quad oldest (s) & 20 \\
\quad optical axes within (deg) & 30 \\
\quad camera distance (m) & 0.3--3 \\
\multicolumn{2}{l}{\emph{A window is refused}} \\
\quad cameras move less than (m) & 0.05 \\
\quad camera centers off by more than (m) & 0.15 \\
\quad log scale of the centers beyond & 0.3 \\
\multicolumn{2}{l}{\emph{Depth filter}} \\
\quad nearest (m) & 0.15 \\
\quad farthest (m) & 20 \\
\quad confidence quantile dropped & 0.2 \\
\quad relative depth step of an edge & 0.15 \\
\quad relative agreement with another image & 0.1 \\
\multicolumn{2}{l}{\emph{Anchor}} \\
\quad older windows triangulated against & 4 \\
\quad shortest baseline (m) & 0.2 \\
\quad longest baseline (m) & 3 \\
\quad least parallax (deg) & 2 \\
\quad largest reprojection error (px) & 2 \\
\quad largest log-depth sigma of a target & 0.1 \\
\quad targets one image of a window needs & 30 \\
\quad $\sigma_0$ & 0.05 \\
\quad bound on $|o_j|$ & 0.3 \\
\quad prior on a window's log scale in the graph & 0.01 \\
\multicolumn{2}{l}{\emph{Graph}} \\
\quad matched points per pair of submaps & 80 \\
\quad point sigma, floor (m) & 0.1 \\
\quad along the ray (fraction of depth) & 0.1 \\
\quad across the ray (fraction of depth) & 0.01 \\
\quad relative log scale of two windows & 0.03 \\
\quad odometry drift (m per m) & 0.005 \\
\quad odometry drift (deg per m) & 0.5 \\
\quad iterations per window & 8 \\
\multicolumn{2}{l}{\emph{Loop closure}} \\
\quad candidates per window & 4 \\
\quad at least this much older (s) & 5 \\
\quad within (m) & 8 \\
\quad inliers at least & 30 \\
\quad reprojection threshold (px) & 2 \\
\quad relative depth agreement & 0.2 \\
\quad disagreement with the odometry at most (sigmas) & 3 \\
\quad moves poses when drift exceeds (loop sigmas) & 1 \\
\quad loop sigma, floor (m) & 0.25 \\
\quad loop sigma per meter of its length & 0.1 \\
\multicolumn{2}{l}{\emph{Agreement between views}} \\
\quad pixel step & 6 \\
\quad viewing rays within (deg) & 30 \\
\quad log ratio beyond which a pixel is left out & 0.5 \\
\quad shared pixels a pair needs & 500 \\
\quad weight of the zero prior (pairs) & 1 \\
\quad Huber threshold (robust sigmas) & 1.5 \\
\multicolumn{2}{l}{\emph{Fusion}} \\
\quad voxel (m) & 0.04 \\
\quad pixel step & 4 \\
 
\bottomrule
\end{tabular}

\end{table}

\paragraph{Context keyframes.} A window's depth call takes at most two earlier keyframes (\cref{sec:context}). A keyframe qualifies if it is 2 to 20\,s older than the window, its optical axis is within $30^\circ$ of the center camera's, and its camera is 0.3 to 3\,m from the center camera (\cref{tab:s_config}). Of those that qualify, the two whose optical axes are closest to the center camera's are taken. A context keyframe is passed with its current pose, so its pose relative to the window carries the filter's drift over at most 20\,s. This drift is small: over a whole sequence, the filter's error is \ovEurocTrajFilter\,cm on the mean of the EuRoC rooms and \resLiveGiantAteOdometryOri\,m on the mean of ORI (\cref{tab:s_traj,tab:s_seq}). The graph models it as 0.5\,cm and $0.5^\circ$ per meter of path (\cref{tab:s_config}) and corrects it after every window.

\paragraph{Cameras and IMUs} of the setups (\cref{tab:setups}). EuRoC: gray $752{\times}480$ at 20\,Hz, IMU at 200\,Hz. TUM-VI: gray fisheye $512{\times}512$ at 20\,Hz, 200\,Hz. ORI: color fisheye $1472{\times}1440$ at 30\,Hz, 1\,kHz. Hilti-Oxford: gray fisheye $720{\times}540$ at 40\,Hz, 400\,Hz. Oxford Spires:
color fisheye $1440{\times}1080$ at 20\,Hz (about 5\,Hz on CC05), 400\,Hz. Phone rig: Note~9 color $640{\times}480$ at \phAndroidCamHz\,Hz, \phAndroidImuHz\,Hz; iPhone color $640{\times}480$ at \phIphoneCamHzMin{} to \phIphoneCamHzMax\,Hz, \phIphoneImuHz\,Hz. GoPro: color, read at $640{\times}360$, \gpFrameHz\,Hz, \gpImuHz\,Hz.

\paragraph{Placing a map.} A map is brought into the scan's frame by the rigid transformation that best aligns the system's trajectory with the reference trajectory, the same one the trajectory error is computed with. No scale is fitted and nothing is fitted to the surface.

\paragraph{Scoring a map.} The part of the scan a map is scored against is fixed per sequence from the reference trajectory and the camera's view alone. Accuracy is
the distance from a map point to the scan, within range of a reference camera. Completeness is the distance to the map from the scan points a reference camera sees. The scan is thinned to one point per centimeter, the map is fused on 4\,cm voxels, and the F-score combines the shares of both distances below 5\,cm. A trajectory more than 10\,m off has failed, and its map counts 0. In the tables, here and in the main text, $\times$ marks a system that failed on every sequence of a dataset. A dash marks a result that does not exist: the system gives no such output, was not run on that setup, or failed on every sequence, so that no error is left to average.

\paragraph{Runs of \method.} The system runs live: once the depth model is loaded, the sequence is played back at its rate, the filter processes every frame, and a window that arrives while the mapping worker is busy is dropped.
ORI, whose camera delivers $1472{\times}1440$ frames at 30\,Hz, is played at 0.3 of its rate, the other public datasets at sensor rate. A sequence's result is the mean of its first three runs.

\paragraph{The other pipelines} run from their released code and configuration on the same camera, changed only where a rig requires it, one run per sequence unless stated. Each map is fused on the same voxels and scored by the same code as ours. \emph{VGGT-SLAM~2.0} is given the undistorted frames. It has no metric scale and takes the scale
of the similarity that fits its camera centers to the reference ($^\dagger$ in the tables and figures, here and in the main text). \emph{ScaRF-SLAM's mapper} runs with its own depth model (DA3-Nested-Giant-Large) on the poses of one of our runs ($^\ddagger$), so it has no trajectory of its own. \emph{MASt3R-Fusion} ships two configurations for one camera with an IMU, for EuRoC and for a hand-held fisheye rig. On ORI, whose equidistant fisheye its loader does not take, it is given the undistorted frames at 10\,Hz as a pinhole camera. Both configurations were run on the development sequence R01 and the better one kept (EuRoC's: \resFusionOriEuroc{} against \resFusionOriHandheld). On Hilti-Oxford the same choice on the development sequence exp14 fell on the
\resHiltiFusionChosen{} one (\resHiltiFusionChoiceHandheld{} against \resHiltiFusionChoiceEuroc), which is also used
on Oxford Spires. \emph{DAVIO} runs from its public release with the same depth model as \method. \emph{FoundationVIO} is described in \cref{sec:s_fvio}.

\paragraph{The IMU's noise.}\label{sec:s_noise} Each rig keeps the noise values of its rig file. We tested whether one rule could set them everywhere (the densities at their documented values and the bias walks at $3\times$ theirs, the rule Hilti-Oxford's rig was given), and one factor at a time around the documented values, on
EuRoC and ORI. On ORI the rule keeps the map ($\nzOriGap$) and every setting scores between \nzOriLo{} and \nzOriHi. On EuRoC it changes the map by $\nzEurocGap$, and the settings range from \nzEurocLo{} to \nzEurocHi. In every setting the filter's displacement against the reference moves by under one percent: the noise changes how
well the map is built, not the scale the filter gives it.

\section{EuRoC and ORI}\label{sec:s_results}
\begin{figure}[t]
  \centering
  \includegraphics[width=\columnwidth]{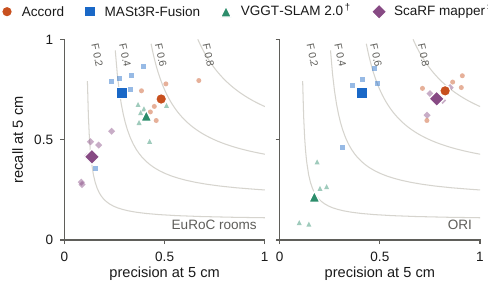}
  \caption{\textbf{Precision and recall at 5\,cm}, the map placed by the system's own trajectory: every sequence
  (small marks) and the mean (large marks); the curves are lines of equal F-score. $^\dagger$, $^\ddagger$ as in
  \cref{tab:s_placement}.}
  \label{fig:s_balance}
\end{figure}

\begin{figure*}[p]
  \centering
  \includegraphics[width=.74\textwidth]{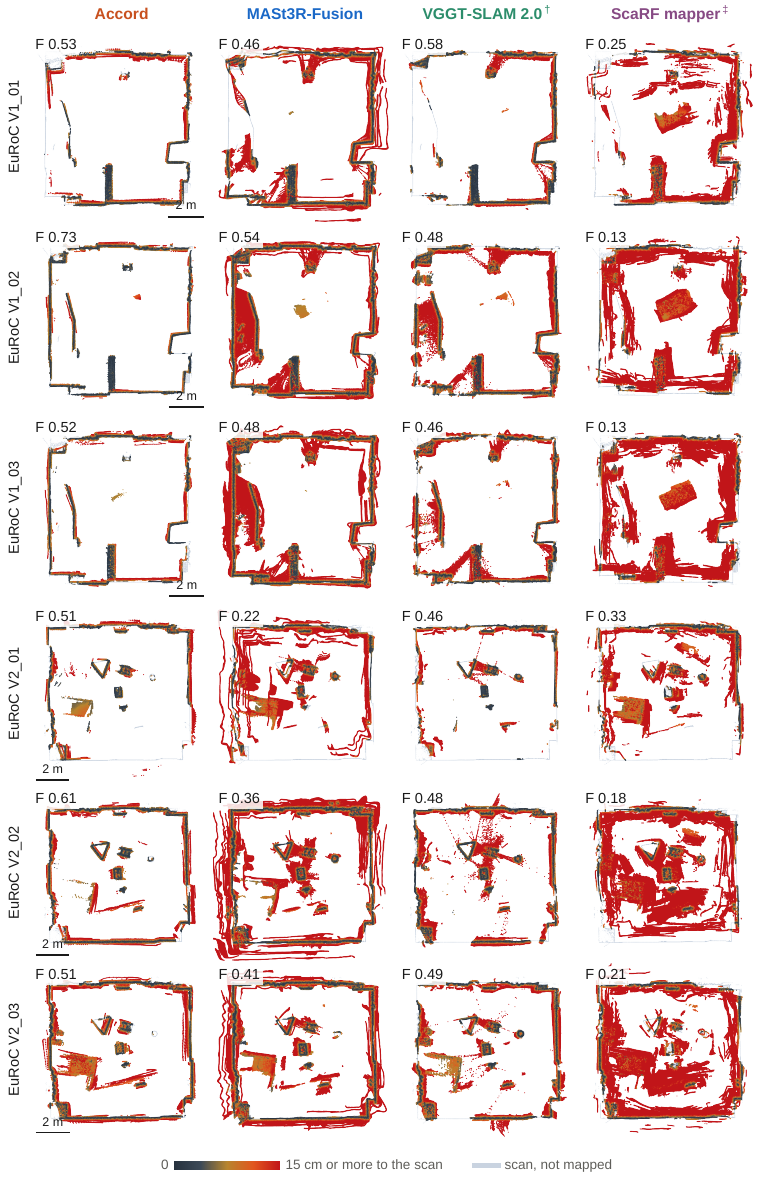}
  \caption{\textbf{The maps of four pipelines on the EuRoC rooms}: a horizontal slab 1.0 to 1.4\,m above the floor, seen from above. Each map is placed by its own trajectory and each of its points is colored by its distance to the scan, which is drawn under the map and shows where nothing was mapped. F: the F-score at 5\,cm of the whole map (\cref{tab:s_maps}). \method: the live run of the median score. $^\dagger$, $^\ddagger$ as in
  \cref{tab:s_placement}.}
  \label{fig:s_maps_rooms}
\end{figure*}

\begin{figure*}[p]
  \centering
  \includegraphics[width=.82\textwidth]{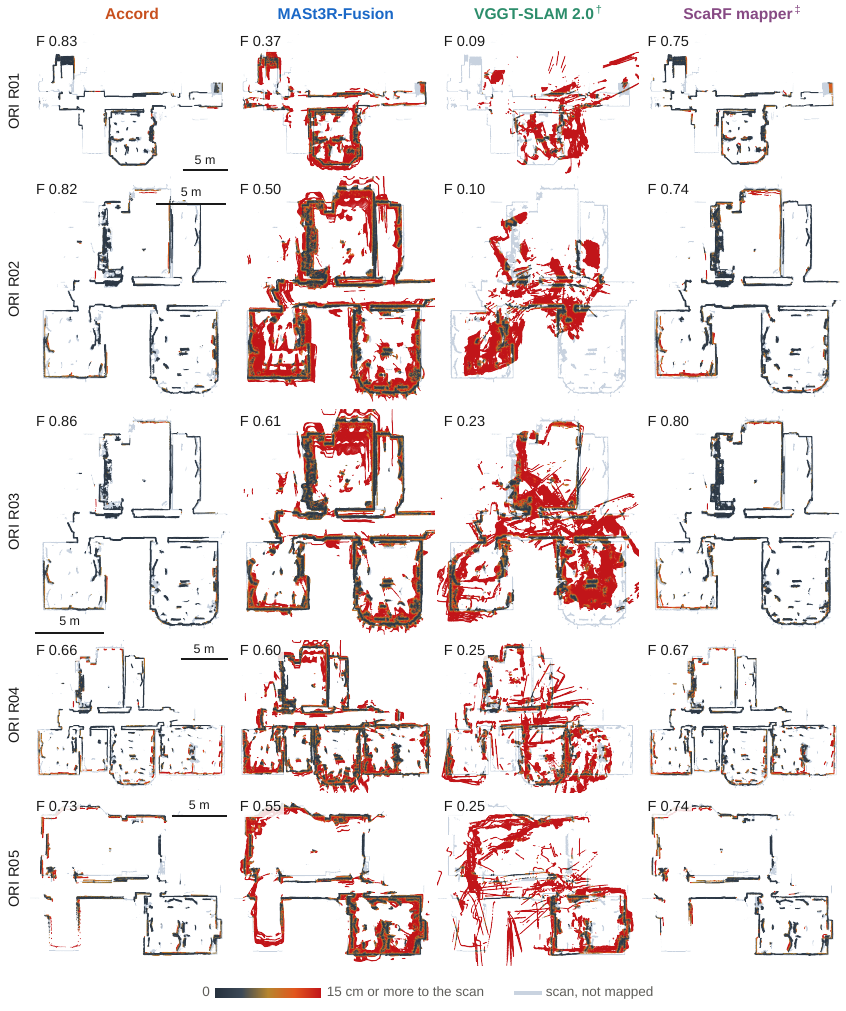}
  \caption{\textbf{The maps of four pipelines on ORI}, drawn as in \cref{fig:s_maps_rooms}.}
  \label{fig:s_maps_ori}
\end{figure*}

\paragraph{Three placements.} \Cref{tab:s_placement} gives the dense maps placed by each system's own trajectory (the protocol of \cref{sec:experiments}), then after a rigid registration to the scan and after a similarity. The order of the
pipelines is the same under all three. \Cref{tab:s_maps} gives every sequence, \cref{fig:s_maps_rooms,fig:s_maps_ori} the maps themselves, and
\cref{fig:s_balance} the precision and recall behind the F-scores: MASt3R-Fusion's maps have the highest mean recall on the rooms and a lower precision than \method's on both datasets.
\begin{table}[t]
  \centering\small
  \caption{\textbf{Dense maps under three placements}, mean F-score at 5\,cm on the \resRoomsN{} EuRoC rooms and the \resOriN{} ORI sequences: placed by the system's own trajectory, after a rigid registration to the scan, and after a similarity. $^\dagger$\,no metric scale: given the reference's. $^\ddagger$\,with its own depth model, on our poses, first placement only. --: not run.}
  \label{tab:s_placement}
  \resizebox{\columnwidth}{!}{\begin{tabular}{@{}lcccccc@{}}
\toprule
 & \multicolumn{3}{c}{EuRoC rooms} & \multicolumn{3}{c}{ORI} \\
\cmidrule(lr){2-4}\cmidrule(l){5-7}
System & traj. & rigid & sim. & traj. & rigid & sim. \\
\midrule
\method, DA3-Giant & \textbf{0.570} & \textbf{0.601} & \textbf{0.632} & \textbf{0.782} & \textbf{0.788} & \textbf{0.852} \\
\method, DA3-Large & 0.521 & 0.561 & 0.588 & 0.698 & 0.711 & 0.798 \\
\method, DA3-Base & 0.499 & 0.539 & 0.562 & 0.677 & 0.693 & 0.751 \\
\midrule
MASt3R-Fusion~\cite{zhou2026mast3rfusion} & 0.411 & 0.434 & 0.442 & 0.525 & 0.533 & 0.556 \\
VGGT-SLAM~2.0~\cite{maggio2026vggtslam2}$^\dagger$ & 0.490 & 0.539 & 0.547 & 0.184 & 0.198 & 0.202 \\
ScaRF-SLAM's mapper~\cite{zhang2026scarfslam}$^\ddagger$ & 0.205 & \na & \na & 0.742 & \na & \na \\
 
\bottomrule
\end{tabular}
}
\end{table}
\begin{table*}[t]
  \centering\small
  \caption{\textbf{The maps of the systems, per sequence}: F-score at 5\,cm placed by the trajectory, after a rigid registration and after a similarity. \method: live, with DA3-Giant. $^\dagger$, $^\ddagger$ as in
  \cref{tab:s_placement}.}
  \label{tab:s_maps}
  \begin{tabular}{@{}lcccccccccc@{}}
\toprule
 & \multicolumn{3}{c}{\method} & \multicolumn{3}{c}{MASt3R-Fusion} & \multicolumn{3}{c}{VGGT-SLAM~2.0$^\dagger$} & ScaRF$^\ddagger$ \\
\cmidrule(lr){2-4}\cmidrule(lr){5-7}\cmidrule(lr){8-10}
Seq. & traj. & rigid & sim. & traj. & rigid & sim. & traj. & rigid & sim. & traj. \\
\midrule
V1\_01 & 0.536 & 0.569 & 0.649 & 0.457 & 0.458 & 0.458 & 0.580 & 0.628 & 0.627 & 0.251 \\
V1\_02 & 0.728 & 0.734 & 0.744 & 0.542 & 0.548 & 0.550 & 0.476 & 0.505 & 0.522 & 0.133 \\
V1\_03 & 0.518 & 0.550 & 0.591 & 0.476 & 0.483 & 0.480 & 0.456 & 0.507 & 0.512 & 0.130 \\
V2\_01 & 0.513 & 0.588 & 0.635 & 0.216 & 0.327 & 0.341 & 0.456 & 0.575 & 0.582 & 0.328 \\
V2\_02 & 0.614 & 0.626 & 0.627 & 0.362 & 0.363 & 0.378 & 0.476 & 0.522 & 0.528 & 0.179 \\
V2\_03 & 0.507 & 0.539 & 0.547 & 0.412 & 0.424 & 0.445 & 0.494 & 0.500 & 0.513 & 0.205 \\
\midrule
R01 & 0.828 & 0.833 & 0.832 & 0.375 & 0.382 & 0.411 & 0.092 & 0.090 & 0.090 & 0.751 \\
R02 & 0.824 & 0.833 & 0.881 & 0.496 & 0.503 & 0.538 & 0.103 & 0.114 & 0.114 & 0.739 \\
R03 & 0.864 & 0.870 & 0.910 & 0.608 & 0.614 & 0.628 & 0.226 & 0.233 & 0.241 & 0.804 \\
R04 & 0.659 & 0.660 & 0.829 & 0.599 & 0.607 & 0.607 & 0.249 & 0.278 & 0.287 & 0.675 \\
R05 & 0.734 & 0.743 & 0.811 & 0.547 & 0.561 & 0.594 & 0.252 & 0.275 & 0.278 & 0.741 \\
 
\bottomrule
\end{tabular}

\end{table*}

\paragraph{Trajectories.} The better map does not come from a better trajectory (\cref{tab:s_traj,tab:overview}). \method's trajectory, corrected by the map, is the best on no dataset: it is off by \ovEurocTrajOurs\,cm in the rooms (the filter as it ran: \ovEurocTrajFilter) against \ovEurocTrajOrb{}
for ORB-SLAM3, by \ovTumviTrajOurs\,cm on TUM-VI (\ovTumviTrajFilter) against \ovTumviTrajOkvis{} for OKVIS2-X after its optimization (\ovTumviTrajOkvisLive{} as it ran), and by \ovHiltiTrajOurs\,cm on Hilti-Oxford (\ovHiltiTrajFilter) against \ovHiltiTrajOkvis{} (\ovHiltiTrajOkvisLive). The map helps the trajectory where the camera returns: loop closure halves the error on ORI, from \resLiveGiantAteOdometryOri{} to
\resLiveGiantAteFinalOri\,m. It does not always help: on Oxford Spires the corrected trajectory is worse on the mean than the filter's (\ovSpiresTrajOurs{} against \ovSpiresTrajOursLive\,cm), from one sequence
(\cref{sec:s_spires}).
\begin{table}[t]
  \centering\small
  \caption{\textbf{Trajectories on the EuRoC rooms}: ATE in meters after a rigid alignment, the median of the runs of the other systems and the mean of \method's. $^\dagger$\,after a similarity, on its keyframes.
  $^\ddagger$\,on its keyframes.}
  \label{tab:s_traj}
  \resizebox{\columnwidth}{!}{\begin{tabular}{@{}lccccccc@{}}
\toprule
 & V1\_01 & V1\_02 & V1\_03 & V2\_01 & V2\_02 & V2\_03 & mean \\
\midrule
ORB-SLAM3~\cite{campos2021orbslam3} & 0.020 & 0.025 & 0.054 & 0.048 & 0.028 & 0.021 & 0.032 \\
OKVIS2~\cite{boche2025okvis2x}, final & 0.034 & 0.021 & 0.040 & 0.131 & 0.055 & 0.077 & 0.060 \\
MASt3R-Fusion, real-time & 0.146 & 0.115 & 0.093 & 0.268 & 0.129 & 0.312 & 0.177 \\
MASt3R-Fusion, global$^\ddagger$ & 0.064 & 0.028 & 0.039 & 0.195 & 0.042 & 0.052 & 0.070 \\
VGGT-SLAM~2.0$^\dagger$ & 0.058 & 0.080 & 0.077 & 0.079 & 0.066 & 0.083 & 0.074 \\
\midrule
\method, the filter & 0.040 & 0.054 & 0.057 & 0.066 & 0.053 & 0.187 & 0.076 \\
\method, corrected by the final map & 0.041 & 0.044 & 0.054 & 0.066 & 0.051 & 0.153 & 0.068 \\
 
\bottomrule
\end{tabular}
}
\end{table}

\paragraph{Our runs per sequence.} \Cref{tab:s_seq} lists the live runs of \method{} per sequence and depth model: the map as built and corrected, under the three placements, the trajectory, and the windows mapped. On ORI the three runs of a sequence give the same map with the giant and with the base model. With the large model some runs dropped windows there, and the scores of a sequence's runs differ by up to \resLiveLargeOriSpread. Against MASt3R-Fusion on ORI our map is ahead on every sequence with each of the three models, narrowly on R04 with the two smaller ones (\resLiveBaseRFour{} and \resLiveLargeRFour{} against its \resOtherFusionRFour).
\begin{table*}[t]
  \centering\small
  \caption{\textbf{\method{} live, per sequence and depth model}: F-score at 5\,cm of the map as built, corrected,  and after a rigid and a similarity registration of the corrected map; ATE of the filter and of the trajectory
  corrected by the final map; windows mapped and not mapped (dropped while the worker was busy, or taken with the camera nearly at rest); loops closed. Means over the runs.}
  \label{tab:s_seq}
  \begin{tabular}{@{}llcccccccccc@{}}
\toprule
 & & & \multicolumn{4}{c}{F-score at 5\,cm} & \multicolumn{2}{c}{ATE (m)} & \multicolumn{2}{c}{Windows} & \\
\cmidrule(lr){4-7}\cmidrule(lr){8-9}\cmidrule(lr){10-11}
Model & Seq. & runs & built & corr. & rigid & sim. & filter & final & mapped & not & loops \\
\midrule
Giant & V1\_01 & 3 & 0.588 & 0.536 & 0.569 & 0.649 & 0.040 & 0.041 & 88.3 & 9.3 & 3.3 \\
Giant & V1\_02 & 3 & 0.658 & 0.728 & 0.734 & 0.744 & 0.054 & 0.044 & 53.0 & 0.3 & 7.0 \\
Giant & V1\_03 & 3 & 0.486 & 0.518 & 0.550 & 0.591 & 0.057 & 0.054 & 66.0 & 0.0 & 6.0 \\
Giant & V2\_01 & 3 & 0.503 & 0.513 & 0.588 & 0.635 & 0.066 & 0.066 & 70.0 & 2.7 & 0.0 \\
Giant & V2\_02 & 3 & 0.510 & 0.614 & 0.626 & 0.627 & 0.053 & 0.051 & 73.3 & 4.3 & 9.3 \\
Giant & V2\_03 & 3 & 0.374 & 0.507 & 0.539 & 0.547 & 0.187 & 0.153 & 73.0 & 0.0 & 1.0 \\
Giant & R01 & 3 & 0.754 & 0.828 & 0.833 & 0.832 & 0.197 & 0.047 & 87.0 & 1.0 & 4.0 \\
Giant & R02 & 3 & 0.743 & 0.824 & 0.833 & 0.881 & 0.124 & 0.062 & 84.0 & 0.0 & 23.0 \\
Giant & R03 & 3 & 0.797 & 0.864 & 0.870 & 0.910 & 0.093 & 0.048 & 80.0 & 1.0 & 39.0 \\
Giant & R04 & 3 & 0.662 & 0.659 & 0.660 & 0.829 & 0.201 & 0.108 & 117.0 & 1.0 & 43.0 \\
Giant & R05 & 3 & 0.717 & 0.734 & 0.743 & 0.811 & 0.095 & 0.094 & 65.0 & 2.0 & 3.0 \\
\midrule
Large & V1\_01 & 3 & 0.565 & 0.560 & 0.583 & 0.612 & 0.040 & 0.042 & 90.3 & 5.3 & 4.7 \\
Large & V1\_02 & 3 & 0.630 & 0.665 & 0.664 & 0.671 & 0.054 & 0.045 & 53.0 & 0.0 & 8.0 \\
Large & V1\_03 & 3 & 0.466 & 0.488 & 0.539 & 0.555 & 0.057 & 0.053 & 66.0 & 0.0 & 6.0 \\
Large & V2\_01 & 3 & 0.503 & 0.463 & 0.535 & 0.603 & 0.066 & 0.066 & 70.0 & 2.0 & 0.0 \\
Large & V2\_02 & 3 & 0.451 & 0.526 & 0.550 & 0.566 & 0.053 & 0.051 & 73.7 & 3.0 & 9.0 \\
Large & V2\_03 & 3 & 0.366 & 0.424 & 0.499 & 0.522 & 0.187 & 0.187 & 73.0 & 0.0 & 0.0 \\
Large & R01 & 3 & 0.759 & 0.801 & 0.807 & 0.809 & 0.197 & 0.049 & 85.7 & 3.7 & 4.0 \\
Large & R02 & 3 & 0.681 & 0.675 & 0.699 & 0.794 & 0.124 & 0.062 & 83.0 & 2.3 & 22.0 \\
Large & R03 & 3 & 0.735 & 0.737 & 0.759 & 0.828 & 0.093 & 0.050 & 76.3 & 9.0 & 36.0 \\
Large & R04 & 3 & 0.631 & 0.604 & 0.610 & 0.767 & 0.201 & 0.109 & 114.0 & 5.0 & 41.0 \\
Large & R05 & 3 & 0.667 & 0.675 & 0.683 & 0.792 & 0.095 & 0.095 & 63.7 & 4.0 & 2.7 \\
\midrule
Base & V1\_01 & 3 & 0.549 & 0.528 & 0.564 & 0.600 & 0.040 & 0.042 & 91.3 & 2.3 & 5.0 \\
Base & V1\_02 & 3 & 0.601 & 0.631 & 0.633 & 0.637 & 0.054 & 0.044 & 53.0 & 0.0 & 8.0 \\
Base & V1\_03 & 3 & 0.459 & 0.464 & 0.503 & 0.503 & 0.057 & 0.053 & 66.0 & 0.0 & 5.0 \\
Base & V2\_01 & 3 & 0.465 & 0.499 & 0.560 & 0.598 & 0.066 & 0.066 & 70.0 & 2.0 & 0.0 \\
Base & V2\_02 & 3 & 0.436 & 0.456 & 0.494 & 0.533 & 0.053 & 0.049 & 74.0 & 1.0 & 12.0 \\
Base & V2\_03 & 3 & 0.348 & 0.417 & 0.480 & 0.503 & 0.187 & 0.189 & 73.0 & 0.0 & 0.0 \\
Base & R01 & 3 & 0.706 & 0.746 & 0.753 & 0.752 & 0.197 & 0.050 & 87.0 & 1.0 & 4.0 \\
Base & R02 & 3 & 0.654 & 0.661 & 0.680 & 0.748 & 0.124 & 0.063 & 84.0 & 0.0 & 21.0 \\
Base & R03 & 3 & 0.740 & 0.734 & 0.756 & 0.799 & 0.093 & 0.048 & 80.0 & 1.0 & 39.0 \\
Base & R04 & 3 & 0.611 & 0.602 & 0.606 & 0.737 & 0.201 & 0.108 & 117.0 & 1.0 & 41.0 \\
Base & R05 & 3 & 0.622 & 0.640 & 0.670 & 0.721 & 0.095 & 0.102 & 65.0 & 2.0 & 3.0 \\
 
\bottomrule
\end{tabular}

\end{table*}

\section{Hilti-Oxford and TUM-VI}\label{sec:s_further}\label{sec:s_hilti}
\paragraph{Sequences.} Hilti-Oxford~\cite{zhang2023hiltioxford} was recorded with a hand-held survey rig in a
historic building (the Sheldonian Theatre) and its surroundings, and on a construction site
(\cref{tab:s_hilti_seqs}). We use its front camera and its IMU with the release's calibration. Every sequence has
control points surveyed to the millimeter, at which the rig stands still; three also have dense reference poses.
Five sequences are for evaluation and were used for no choice; choices were made on five development sequences,
and on two more from the construction site for the update at rest below. The scans of the building match its
sequences. Those of the construction site do not match the sequences recorded there, so maps are scored in the
building only: the three evaluation sequences exp18, exp16 and exp21 are the ``Hilti-Oxford'' map column of
\cref{tab:overview}.
\begin{table}[t]
  \centering\small
  \caption{\textbf{The Hilti-Oxford sequences} we use. Reference: dense poses and control points, or control
  points only. Scan: whether the map is scored (yes), or why not: the site's scan does not match the sequence
  (no), there is no scan (none), or the sequence was run with the filter alone and has no map (--).}
  \label{tab:s_hilti_seqs}
  \resizebox{\columnwidth}{!}{\begin{tabular}{@{}llllc@{}}
  \toprule
  & Sequence & Site & Reference & Scan \\
  \midrule
  \multirow{7}{*}{\rotatebox{90}{development}}
  & exp14 Basement 2 & building & dense & yes \\
  & exp11 Lower Gallery & building & points & yes \\
  & exp15 Attic to Upper Gallery & building & points & yes \\
  & exp05 Construction Upper Level 2 & site & points & no \\
  & exp07 Long Corridor & site & points & none \\
  & exp02 Construction Multilevel & site & points & -- \\
  & exp06 Construction Upper Level 3 & site & points & -- \\
  \midrule
  \multirow{5}{*}{\rotatebox{90}{evaluation}}
  & exp18 Corridor Lower Gallery 2 & building & dense & yes \\
  & exp16 Attic to Upper Gallery 2 & building & dense & yes \\
  & exp21 Outside Building & building & points & yes \\
  & exp04 Construction Upper Level & site & points & no \\
  & exp01 Construction Ground Level & site & points & no \\
  \bottomrule
  \end{tabular}}
\end{table}

\paragraph{Scoring.} Where the reference is dense, a trajectory is scored by its ATE after a rigid alignment. At
the control points it is scored by the dataset's own protocol: one rigid fit to the points it covers, the RMSE
of their errors, and the dataset's score, in which a point scores 10, 6, 3, 1 or 0 for an error below 1, 3, 6 or
10\,cm or above, given in percent of its maximum. Maps are scored as on EuRoC and ORI (\cref{sec:s_protocol}); where the reference is control points
only, the part of the scan a map is scored against is the one the cameras of our run see, the same for every
system.

\paragraph{Two settings of the filter.} Hilti-Oxford is the one platform where the filter needed more than its
rig file. Both settings were chosen on the development sequences, with the filter alone, by rules written before
the runs.

\emph{The IMU's noise.} With the noise of the dataset's first configuration, five times the sensor's data sheet,
the filter's scale was too long by up to 14\,\% on the development sequences. The release ships an hour and a
half of the IMU at rest; its Allan deviation gives the white noise of the accelerometer and the gyroscope and
upper bounds on their random walks. Among the densities at one to three times these values and the walks at one
to ten times their bounds, the setting with the lowest mean error at the control points was taken
(\cref{tab:s_hilti_noise}): the densities as measured and the walks at three times their bounds.
\begin{table}[t]
  \centering\small
  \caption{\textbf{The IMU's noise on Hilti-Oxford}, the filter alone: mean RMSE (m) at the control points of the
  five development sequences.}
  \label{tab:s_hilti_noise}
  \resizebox{\columnwidth}{!}{\begin{tabular}{@{}lc@{}}
  \toprule
  Noise & RMSE \\
  \midrule
  the dataset's first configuration & 0.627 \\
  densities as measured, $\times$1 / 1.5 / 2 / 3 & 0.167 / 0.201 / 0.296 / 0.415 \\
  and walks at their bounds, $\times$1 / 3 / 10 & 0.154 / \textbf{0.159} / 0.173 \\
  \bottomrule
  \end{tabular}}
\end{table}

\emph{An update at rest.} Every sequence stops at each control point. While the rig stands, the features the
filter tracks have no parallax, and the landmarks already in its state hold the pose. That held in 22 of 23 stands
of the development sequences. In the other, on exp07, the camera stands before a blank wall at close range, the
landmarks leave the state within a frame, and the filter, on the IMU alone, drifts 0.57\,m and jumps 0.22\,m when
the walk resumes. We therefore update the filter while the rig is at rest, without dropping the camera frame as
OpenVINS's own zero-velocity update does: while the gyroscope and the accelerometer are quiet over the last half
second, the filter is given the IMU's measurement at rest (no rotation, gravity in the current attitude) and then
a zero velocity, both through OpenVINS's own update. On the seven development sequences this raises the mean
dataset score from 6.04 to 6.53 and lowers the mean RMSE at the control points from 0.874 to 0.362\,m. It is on
for Hilti-Oxford only; the system's configuration is unchanged elsewhere.

\paragraph{Where the system fails.} The update at rest removes the jump on exp07 (0.22 to 0.04\,m) but not the
drift during the stand. When the landmarks leave, the filter's velocity is already 0.15\,m/s off while its
covariance gives it a sigma of 0.8\,mm/s: the filter is far too sure of its velocity, so each update removes only
a small part of the error. On the evaluation sequences of the construction site the filter's error at the control
points therefore stays large, \resHiltiFourFinal{} and \resHiltiOneFinal\,m on exp04 and exp01. On exp11 the trajectory
corrected by the final map is worse than the filter's (\resHiltiElevenFinal{} against \resHiltiElevenOdometry\,m),
where one loop closes; we have not found why. Of the other systems, ORB-SLAM3 did not initialize on exp04,
MASt3R-Fusion's global trajectory diverged on exp16, and VGGT-SLAM~2.0, scaled to the reference, is off by a meter
or more on every sequence.

\paragraph{Every sequence.} \Cref{tab:s_hilti_traj,tab:s_hilti_maps} give every system on every sequence, \method{}
with DA3-Base beside DA3-Giant. Our three runs of a sequence repeat exactly on eight of the ten
sequences.
\begin{table*}[t]
  \centering\small
  \caption{\textbf{Trajectories on Hilti-Oxford}: the error in meters (ATE where the reference is dense, exp14,
  exp18 and exp16; the RMSE at the control points elsewhere) and the dataset's score. fail: no run initialized;
  div.: more than 10\,m off (failed). $^\dagger$\,given the reference's scale.}
  \label{tab:s_hilti_traj}
  \resizebox{\textwidth}{!}{\begin{tabular}{@{}lcccccccccc@{}}
  \toprule
  & \multicolumn{5}{c}{development} & \multicolumn{5}{c}{evaluation} \\
  \cmidrule(lr){2-6}\cmidrule(l){7-11}
  System & exp14 & exp05 & exp07 & exp11 & exp15 & exp18 & exp16 & exp04 & exp01 & exp21 \\
  \midrule
  ORB-SLAM3~\cite{campos2021orbslam3} & fail & 0.170 / 1.7 & fail & 0.291 / 3.0 & 0.234 / 2.2 & 0.127 & 0.157 / 3.3 & fail & 0.948 / 0.0 & 1.495 / 0.0 \\
OKVIS2~\cite{boche2025okvis2x}, final & 0.054 / 27.5 & 0.162 / 5.0 & 0.589 / 0.0 & 0.147 / 6.0 & 0.224 / 0.0 & 0.205 / 20.0 & 0.229 / 3.3 & 0.108 / 22.9 & 0.104 / 15.4 & 0.210 / 0.0 \\
MASt3R-Fusion~\cite{zhou2026mast3rfusion} & 0.178 / 0.0 & 0.368 / 0.0 & 0.808 / 0.0 & 0.120 / 8.0 & 0.328 / 0.0 & 0.139 & div. & 0.458 / 1.4 & 1.578 / 0.0 & 1.174 / 0.0 \\
VGGT-SLAM~2.0~\cite{maggio2026vggtslam2}$^\dagger$ & 1.009 / 0.0 & 4.737 / 0.0 & div. & 2.592 / 0.0 & 6.631 / 0.0 & 6.464 & 4.647 / 0.0 & 4.689 / 0.0 & 3.690 / 0.0 & 2.424 / 0.0 \\
OKVIS2, live & 0.093 / 20.0 & 0.173 / 1.7 & 0.785 / 0.0 & 0.230 / 4.0 & 0.305 / 0.0 & 0.230 / 20.0 & 0.322 / 0.0 & 0.195 / 4.3 & 0.160 / 10.0 & 0.464 / 0.0 \\
MASt3R-Fusion, real-time & 0.382 / 30.0 & 0.418 / 0.0 & 0.844 / 0.0 & 0.136 / 0.0 & 0.418 / 0.0 & 0.302 & div. & 0.707 / 1.4 & 2.294 / 0.0 & 1.607 / 0.0 \\
\midrule
\method, DA3-Base & 0.052 / 12.5 & 0.087 / 15.0 & 0.219 / 0.0 & 0.157 / 4.0 & 0.194 / 1.1 & 0.104 & 0.158 / 5.6 & 0.314 / 0.0 & 0.557 / 0.8 & 0.806 / 0.0 \\
\method, DA3-Giant & 0.052 / 12.5 & 0.084 / 20.0 & 0.219 / 0.0 & 0.160 / 4.0 & 0.203 / 0.0 & 0.106 & 0.154 / 7.8 & 0.281 / 0.0 & 0.551 / 0.8 & 0.769 / 0.0 \\
 
  \bottomrule
  \end{tabular}}
\end{table*}
\begin{table*}[t]
  \centering\small
  \caption{\textbf{Maps on Hilti-Oxford}, the sequences whose scan matches them: F-score at 5\,cm placed by the
  system's own trajectory, and after a rigid registration to the scan. div.: the trajectory more than 10\,m off
  (failed). $^\dagger$, $^\ddagger$ as in \cref{tab:s_placement}.}
  \label{tab:s_hilti_maps}
  \resizebox{\textwidth}{!}{\begin{tabular}{@{}lcccccccccccc@{}}
  \toprule
  & \multicolumn{6}{c}{development} & \multicolumn{6}{c}{evaluation} \\
  \cmidrule(lr){2-7}\cmidrule(l){8-13}
  & \multicolumn{2}{c}{exp14} & \multicolumn{2}{c}{exp11} & \multicolumn{2}{c}{exp15}
  & \multicolumn{2}{c}{exp18} & \multicolumn{2}{c}{exp16} & \multicolumn{2}{c}{exp21} \\
  System & traj. & rigid & traj. & rigid & traj. & rigid & traj. & rigid & traj. & rigid & traj. & rigid \\
  \midrule
  MASt3R-Fusion~\cite{zhou2026mast3rfusion} & 0.313 & 0.409 & 0.180 & 0.281 & 0.183 & 0.199 & 0.288 & 0.299 & div. & div. & 0.093 & 0.186 \\
VGGT-SLAM~2.0~\cite{maggio2026vggtslam2}$^\dagger$ & 0.076 & 0.088 & 0.064 & 0.064 & 0.031 & 0.034 & 0.032 & 0.033 & 0.028 & 0.028 & 0.057 & 0.064 \\
ScaRF-SLAM's mapper~\cite{zhang2026scarfslam}$^\ddagger$ & 0.209 & 0.242 & 0.153 & 0.157 & 0.073 & 0.129 & 0.120 & 0.129 & 0.107 & 0.112 & 0.053 & 0.138 \\
\midrule
\method, DA3-Base & 0.384 & 0.398 & 0.205 & 0.277 & 0.174 & 0.232 & 0.176 & 0.213 & 0.179 & 0.181 & 0.066 & 0.147 \\
\method, DA3-Giant & 0.474 & 0.501 & 0.228 & 0.360 & 0.183 & 0.239 & 0.218 & 0.248 & 0.198 & 0.230 & 0.075 & 0.154 \\
 
  \bottomrule
  \end{tabular}}
\end{table*}

\paragraph{TUM-VI.} TUM-VI~\cite{schubert2018tumvi}: the six rooms, a hand-held fisheye under motion capture
throughout; there is no scan, so only trajectories are scored (\cref{tab:s_tumvi}). Our system runs as on EuRoC,
with its configuration unchanged, three times per room; nothing was chosen on these sequences.
\begin{table*}[t]
  \centering\small
  \caption{\textbf{Trajectories on TUM-VI}: ATE in meters after a rigid alignment, the median of the runs of the
  other systems and the mean of \method's. fail: no pose. $^\dagger$\,after a similarity, on its keyframes.
  $^\S$\,on its keyframes.}
  \label{tab:s_tumvi}
  \begin{tabular}{@{}lccccccc@{}}
  \toprule
  System & room1 & room2 & room3 & room4 & room5 & room6 & mean \\
  \midrule
  ORB-SLAM3~\cite{campos2021orbslam3} & 0.021 & 0.018 & 0.040 & 0.018 & 0.013 & 0.007 & 0.020 \\
OKVIS2~\cite{boche2025okvis2x}, final & 0.012 & 0.014 & 0.012 & 0.013 & 0.013 & 0.021 & 0.014 \\
OKVIS2, live & 0.024 & 0.019 & 0.057 & 0.023 & 0.281 & 0.033 & 0.073 \\
MASt3R-Fusion, real-time & 0.168 & 0.519 & 0.513 & 0.139 & 0.196 & 0.313 & 0.308 \\
MASt3R-Fusion, global$^\S$ & 0.020 & 0.037 & 0.041 & 0.022 & 0.019 & 0.041 & 0.030 \\
VGGT-SLAM~2.0$^\dagger$ & 0.247 & 0.268 & 0.467 & 0.126 & 0.209 & 0.182 & 0.250 \\
\midrule
\method, the filter & 0.063 & 0.072 & 0.096 & 0.042 & 0.072 & 0.065 & 0.068 \\
\method, corrected by the final map & 0.037 & 0.027 & 0.049 & 0.042 & 0.051 & 0.065 & 0.045 \\
 
  \bottomrule
  \end{tabular}
\end{table*}

\section{Oxford Spires}\label{sec:s_spires}
\paragraph{Setup.} Oxford Spires~\cite{tao2025spires} was recorded with a rig carried on foot through Oxford
colleges and laser-scanned with a survey-grade scanner. We use its front camera and its IMU on the three Christ
Church sequences that FoundationVIO's paper evaluates on, CC02, CC03 and CC05; nothing was chosen on them. CC02
and CC05 cross the quads, whose far sides lie beyond the depth model's range of 20\,m. The rig file comes from the
dataset's calibration only, and the system runs with its one configuration, three runs per sequence at sensor
rate. The reference is the dataset's trajectory, registered to the scan, and its 1\,cm merged scan.

\paragraph{Against FoundationVIO.} \Cref{tab:s_spires} compares the two systems on each sequence. \method's map is ahead on every sequence, \spOursMap{}
against \spFvioMap{} on the mean, and ahead on both precision and recall on CC02 and CC05. On CC03 FoundationVIO's
points are the closer ones (its precision is higher) while \method's cover about twice as much of the scan.
FoundationVIO's map is a band along its path, where its depth is kept; \method's reaches the facades across the
quads (\cref{fig:spires}). \method's trajectory, corrected by the map, is off by \spOursTraj{}
against \spFvioTraj\,m, ahead on \spAheadTraj{} of 3; on CC02 loop closure makes it worse than the filter's own.
\begin{table}[t]
  \centering\small
  \caption{\textbf{Oxford Spires, against FoundationVIO}: F-score at 5\,cm of the map placed by its own
  trajectory, with its precision (P) and recall (R); \method{} as built and corrected (means of three runs),
  FoundationVIO's one run. Trajectory error (m) after a rigid fit: \method's corrected by the map and as the filter
  published it, FoundationVIO's as it outputs it.}
  \label{tab:s_spires}
  \resizebox{\columnwidth}{!}{%
  \begin{tabular}{@{}lcccc@{\hspace{8pt}}ccc@{\hspace{8pt}}ccc@{}}
  \toprule
  & \multicolumn{4}{c}{\method: map} & \multicolumn{3}{c}{FoundationVIO: map} & \multicolumn{3}{c}{Trajectory (m)} \\
  \cmidrule(r{8pt}){2-5}\cmidrule(r{8pt}){6-8}\cmidrule{9-11}
  Seq. & built & corrected & P & R & F & P & R & \method & filter & FVIO \\
  \midrule
  CC02 & 0.131 & 0.137 & 0.16 & 0.12 & 0.082 & 0.13 & 0.06 & 1.09 & 0.63 & 1.13 \\
CC03 & 0.242 & 0.265 & 0.25 & 0.29 & 0.191 & 0.28 & 0.15 & 0.30 & 0.39 & 0.16 \\
CC05 & 0.185 & 0.187 & 0.21 & 0.17 & 0.074 & 0.11 & 0.06 & 0.80 & 0.82 & 1.86 \\
\midrule
Mean & 0.186 & 0.196 & 0.21 & 0.19 & 0.116 & 0.17 & 0.09 & 0.73 & 0.61 & 1.05 \\
 
  \bottomrule
  \end{tabular}}
\end{table}

\paragraph{Every dense pipeline.} \Cref{tab:s_spires_all} gives every dense pipeline, each configured as on
Hilti-Oxford (\cref{sec:s_protocol}). \method{} has the best map on every sequence. MASt3R-Fusion's trajectory fails on every
sequence and VGGT-SLAM~2.0's on CC05; DAVIO as released makes no map on CC02 and CC03 in any of its three runs, and
on CC05 its trajectory fails.
\begin{table}[t]
  \centering\small
  \caption{\textbf{Every dense pipeline on Oxford Spires}: the F-score at 5\,cm of the map placed by its own
  trajectory, and the trajectory's error (m). \method{} and DAVIO: means of three runs; the others: one run. --: no
  map, or the map of a failed trajectory; failed: a trajectory more than 10\,m off. In the means, a map not made
  or of a failed trajectory counts 0 ($^{n}$: $n$ such), and a failed trajectory is left out ($^{n}$: $n$ left
  out). $^\dagger$\,given the reference's scale. $^\ddagger$\,on our poses, so no trajectory of its own.}
  \label{tab:s_spires_all}
  \resizebox{\columnwidth}{!}{%
  \begin{tabular}{@{}lcccc@{\hspace{8pt}}cccc@{}}
  \toprule
  & \multicolumn{4}{c}{Map: F-score at 5\,cm $\uparrow$} & \multicolumn{4}{c}{Trajectory (m) $\downarrow$} \\
  \cmidrule(r{8pt}){2-5}\cmidrule{6-9}
  & CC02 & CC03 & CC05 & mean & CC02 & CC03 & CC05 & mean \\
  \midrule
  MASt3R-Fusion & -- & -- & -- & 0.000$^{3}$ & failed & failed & failed & --$^{3}$ \\
VGGT-SLAM~2.0$^\dagger$ & 0.042 & 0.096 & -- & 0.046$^{1}$ & 7.12 & 0.53 & failed & 3.82$^{1}$ \\
ScaRF mapper$^\ddagger$ & 0.044 & 0.063 & 0.041 & 0.049 & \na & \na & \na & \na \\
FoundationVIO & 0.082 & 0.191 & 0.074 & 0.116 & 1.13 & \textbf{0.16} & 1.86 & 1.05 \\
DAVIO, as released & -- & -- & -- & 0.000$^{3}$ & \textbf{0.87} & 0.52 & failed & 0.70$^{1}$ \\
\midrule
\method & \textbf{0.137} & \textbf{0.265} & \textbf{0.187} & \textbf{0.196} & 1.09 & 0.30 & \textbf{0.80} & \textbf{0.73} \\
 
  \bottomrule
  \end{tabular}}
\end{table}

\section{FoundationVIO against its own paper}\label{sec:s_fvio}
\paragraph{How it was run.} We ran FoundationVIO's release (commit a4148a6) with its own configurations
(DA3-Nested-Giant-Large~1.1, images at 5\,Hz in batches of 10, loop closure on), one run per sequence, on an
RTX~4090. Its map is its per-image clouds merged into 4\,cm voxels, scored as ours.

\paragraph{Two protocols.} FoundationVIO's paper scores ORI under the protocol of ScaRF-SLAM: the path is cut into
10\,m pieces and each piece is fitted and registered to the scan before it is scored, which forgives drift.
\Cref{tab:s_fvio} therefore scores both systems under that protocol as well, with ScaRF-SLAM's
\texttt{compare\_pts.py} as released and with the rule FoundationVIO's paper states for recall (the scan within
30\,cm of the map only). Under that rule, on the mean of the five sequences, FoundationVIO's paper reports
\fvioScarfPaper\,\%, our run of its release reaches \fvioScarfRunCut\,\% and \method{} \oursScarfCut\,\%:
these are the three figures of \cref{sec:exp_map}. \method{} is ahead of the figures FoundationVIO's paper reports on
\oursScarfCutAheadPaper{} of 5 sequences and of our run of it on \oursScarfCutAheadRun. Under our protocol,
placed by its own trajectory, FoundationVIO scores \fvioOriMap{} on ORI.

\paragraph{FoundationVIO on EuRoC.} On the \fvioEurocRooms{} EuRoC rooms all \fvioEurocDiverged{} runs diverge: its initialization fixes the scale while the drone has barely moved, and its trajectory ends \fvioEurocAteMinKm{} to \fvioEurocAteMaxKm\,km off. It ships no configuration for EuRoC and we wrote none for it.
\begin{table}[t]
  \centering\small
  \caption{\textbf{FoundationVIO against its paper} on ORI. F-score at 5\,cm (\%) under the protocol of its paper
  (ScaRF-SLAM's \texttt{compare\_pts.py}): as the tool is released, and with the rule its paper states for recall.
  \method: the corrected map. APE (m) after a rigid fit: as its paper reports it, and our run scored by its own
  evaluation script.}
  \label{tab:s_fvio}
  \resizebox{\columnwidth}{!}{%
  \begin{tabular}{@{}lccccc@{\hspace{8pt}}cc@{}}
  \toprule
  & \multicolumn{5}{c}{F-score at 5\,cm (\%), its protocol} & \multicolumn{2}{c}{APE (m)} \\
  \cmidrule(r{8pt}){2-6}\cmidrule{7-8}
  & \multicolumn{3}{c}{FoundationVIO} & \multicolumn{2}{c}{\method} & \multicolumn{2}{c}{FoundationVIO} \\
  \cmidrule(r){2-4}\cmidrule(r{8pt}){5-6}\cmidrule{7-8}
  Seq. & its paper & our run: tool & our run: its rule & tool & its rule & its paper & our run \\
  \midrule
  R01 & 76.2 & 63.1 & 67.9 & 72.8 & 77.4 & 0.188 & 0.272 \\
R02 & 74.4 & 61.0 & 65.4 & 66.1 & 71.9 & 0.070 & 0.093 \\
R03 & 75.7 & 61.8 & 65.4 & 72.9 & 77.2 & 0.064 & 0.128 \\
R04 & 72.6 & 53.7 & 57.1 & 70.5 & 74.3 & 0.127 & 0.255 \\
R05 & 66.6 & 61.8 & 66.9 & 67.6 & 70.6 & 0.058 & 0.063 \\
\midrule
Mean & 73.1 & 60.3 & 64.5 & 70.0 & 74.3 & 0.101 & 0.162 \\
 
  \bottomrule
  \end{tabular}}
\end{table}

\section{Everyday devices}\label{sec:s_devices}
This section describes the phone rig dataset of \cref{sec:exp_devices} (how it was recorded, how
its reference is made and how far it can be trusted, and every recording's scores) and the GoPro walk. The phone
recordings, the reference, the rig files and the scripts that made them will be released.

\begin{table*}[t]
  \centering\small
  \caption{\textbf{The walks of the phone rig dataset.} Time and path are the reference's; closures: RTAB-Map's
  loop closures; each phone's span is its recording's (the Note~9 has none of VIO03, whose camera stopped after a
  few seconds; the iPhone stopped early on most walks); ARKit: the iPhone's own tracking against the reference
  after a rigid fit, over the iPhone's part of the walk. Below the rule, the walks added after the first
  \phFirstN{} had been benchmarked.}
  \label{tab:s_phone_walks}
  \setlength{\tabcolsep}{3pt}
\begin{tabular}{@{}llcccccc@{}}
\toprule
Walk & Scene & Time & Path (m) & Closures & Note~9 (s) & iPhone (s) & ARKit (m) \\
\midrule
VIO01 & indoor: corridors of a building & 3:36 & 128 & 28 & 221 & 120 & 0.15 \\
VIO02 & indoor: corridors and a stairwell over several floors & 6:32 & 249 & 44 & 394 & 130 & 0.17 \\
VIO03 & outdoor: paved courtyard with bicycle racks & 3:58 & 180 & 10 & -- & 238 & 0.23 \\
VIO04 & outdoor: courtyard with planters and steps & 5:13 & 223 & 3 & 316 & 120 & 0.14 \\
VIO05 & outdoor: road around a block, parked cars & 4:45 & 283 & 41 & 286 & 238 & 0.44 \\
VIO06 & outdoor: park with playground equipment & 6:30 & 294 & 58 & 396 & 120 & 0.14 \\
VIO07 & outdoor: walled oval field, grass and bare ground & 4:08 & 210 & 44 & 251 & 120 & 0.10 \\
VIO08 & outdoor: sports ground, running track and pitch & 6:10 & 354 & 11 & 374 & 120 & 0.19 \\
VIO11 & outdoor: playground on gravel & 3:30 & 111 & 17 & 212 & 215 & 0.25 \\
\midrule
VIO12 & night: an enclosed courtyard under lamps & 9:21 & 285 & 44 & 547 & 120 & 0.16 \\
VIO13 & night: the same courtyard, another route & 5:18 & 188 & 13 & 323 & 269 & 0.21 \\
 
\bottomrule
\end{tabular}

\end{table*}

\paragraph{Recording.} Three phones were carried together by hand through a university campus on \phWalksN{}
walks: \phIndoorN{} indoors (corridors, and a stairwell over several floors), \phOutdoorN{} outdoors by day and
\phNightN{} at night under lamps, \phMinutes{} minutes and \phKm\,km in all (\cref{tab:s_phone_walks}). The phones
are a Samsung Galaxy Note~9, an iPhone, and a second, LiDAR-equipped iPhone running
RTAB-Map~\cite{labbe2019rtabmap}, whose result is the reference. Each phone records on its own clock. The walks
are at a walking pace and do not start from rest.

\paragraph{The rig files.}\label{sec:s_phone_rig} No calibration target and no calibration recording were used.
The Note~9's focal length comes from the phone's metadata and the iPhone's intrinsics are the median of ARKit's
estimates, both without a distortion model. The camera's rotation on the IMU is the device's own axes, checked
against the gyroscope, with a zero lever arm, and the IMU's noise is assumed. The phones' clocks are aligned from their gyroscopes, whose rates correlate at
\phGyroCorrMin{} or better on every walk, and the Note~9's frames are shifted by one value for the device, since
Android stamps a frame at the start of its exposure. Each phone's rotation to the others is one value for the
dataset. The lever arms were not measured, so each phone's reference keeps the LiDAR iPhone's position and no
error smaller than the rig means anything; the one exception is the iPhone's height: it rode
\phIphoneBelowCm\,cm below the LiDAR iPhone (measured from its own LiDAR), and its reference is lowered by that,
for every system alike.

\paragraph{The reference and its limits.}\label{sec:s_phone_ref} RTAB-Map on the LiDAR iPhone, with ARKit's
odometry, the LiDAR and its own loop closures, gives a graph of
about one pose a second and a colored map. It is the best estimate available, not ground truth in the survey sense. 

\paragraph{Privacy.} People, faces and number plates were detected on every frame at low thresholds, linked over
neighboring frames and blurred. Every system runs on the released, blurred frames, so its numbers can be reproduced from the release.

\paragraph{The benchmark.} Every system runs with its configuration of the public setups and the phone's rig file. \method{} and DAVIO are run three times per recording and reported by the
mean, ORB-SLAM3 and OKVIS2-X three times and by the median, the others once. A trajectory more than \phFailM\,m off, or none, has failed: it is left out of the ATE means and its map counts 0.

\paragraph{Results per recording.} \Cref{tab:s_phone_maps} gives the map of every recording
(\cref{tab:devices} gives the means). \method's map has the highest score
on \phAndroidLead{} of the \phAndroidN{} Note~9 recordings and on \phIphoneLead{} of the \phIphoneN{} iPhone
recordings. Our filter fails on three recordings (the Note~9's VIO02, and the iPhone's VIO06 and VIO12, the
latter at night), where FoundationVIO fails on none. On the recordings both track, \method's trajectory error is the
lower one: \phAndroidCommonFvioOurs{} against \phAndroidCommonFvioOther\,m on the Note~9 and
\phIphoneCommonFvioOurs{} against \phIphoneCommonFvioOther\,m on the iPhone.
\begin{table*}[p]
  \centering\small
  \caption{\textbf{The phone rig, maps}: per recording, the F-score at 5\,cm against RTAB-Map's map, placed by the
  system's own trajectory, points beyond 3\,m of the reference track left out (their share, in \%, below each
  phone). 0.00: no map, or the map of a failed trajectory. \method{} and DAVIO: the mean of three runs. Bold: the
  best of the recording. $^\dagger$\,given the reference's scale; $^\ddagger$\,on our poses.}
  \label{tab:s_phone_maps}
  \setlength{\tabcolsep}{4pt}
\begin{tabular}{@{}lcccccc@{}}
\toprule
Recording & MASt3R-F. & VGGT-SLAM$^\dagger$ & ScaRF$^\ddagger$ & FVIO & DAVIO & \method \\
\midrule
\multicolumn{7}{l}{\emph{Note~9}} \\
VIO01 & 0.19 & 0.06 & 0.21 & 0.17 & 0.26 & \textbf{0.41} \\
VIO02 & 0.00 & 0.02 & 0.00 & \textbf{0.11} & 0.00 & 0.00 \\
VIO04 & 0.00 & 0.15 & 0.17 & 0.08 & 0.14 & \textbf{0.26} \\
VIO05 & 0.08 & 0.06 & 0.08 & 0.07 & 0.08 & \textbf{0.11} \\
VIO06 & 0.00 & 0.15 & 0.17 & 0.19 & 0.26 & \textbf{0.30} \\
VIO07 & 0.00 & 0.08 & 0.18 & \textbf{0.19} & 0.15 & \textbf{0.19} \\
VIO08 & 0.00 & 0.00 & 0.06 & \textbf{0.07} & 0.00 & 0.05 \\
VIO11 & 0.17 & 0.10 & 0.23 & 0.23 & 0.00 & \textbf{0.39} \\
VIO12 & 0.00 & 0.09 & 0.16 & \textbf{0.19} & \textbf{0.19} & \textbf{0.19} \\
VIO13 & 0.00 & 0.02 & 0.14 & \textbf{0.21} & 0.11 & 0.20 \\
\midrule Mean, first nine walks & 0.05$^{5}$ & 0.08$^{1}$ & 0.14$^{1}$ & 0.14 & 0.11$^{3}$ & 0.21$^{1}$ \\
\midrule Mean, all walks & 0.04$^{7}$ & 0.07$^{1}$ & 0.14$^{1}$ & 0.15 & 0.12$^{3}$ & 0.21$^{1}$ \\
Beyond 3\,m, \% & 41 & 31 & 8 & 17 & 13 & 15 \\
\midrule
\multicolumn{7}{l}{\emph{iPhone}} \\
VIO01 & 0.00 & 0.02 & 0.13 & \textbf{0.33} & 0.13 & 0.14 \\
VIO02 & 0.00 & 0.01 & 0.06 & \textbf{0.12} & 0.00 & 0.05 \\
VIO03 & 0.00 & 0.18 & 0.15 & \textbf{0.24} & 0.03 & 0.08 \\
VIO04 & 0.00 & 0.11 & 0.13 & 0.05 & 0.15 & \textbf{0.16} \\
VIO05 & 0.00 & 0.00 & \textbf{0.10} & 0.01 & 0.09 & \textbf{0.10} \\
VIO06 & 0.00 & 0.06 & 0.00 & \textbf{0.07} & 0.00 & 0.00 \\
VIO07 & 0.00 & 0.21 & 0.02 & 0.15 & \textbf{0.24} & 0.06 \\
VIO08 & 0.00 & 0.00 & 0.12 & 0.07 & 0.00 & \textbf{0.25} \\
VIO11 & 0.00 & 0.15 & \textbf{0.33} & 0.19 & 0.29 & 0.32 \\
VIO12 & 0.00 & \textbf{0.14} & 0.00 & 0.06 & 0.11 & 0.00 \\
VIO13 & 0.00 & 0.14 & 0.12 & 0.11 & 0.12 & \textbf{0.15} \\
\midrule Mean, first nine walks & 0.00$^{9}$ & 0.08$^{1}$ & 0.11$^{1}$ & 0.14 & 0.10$^{3}$ & 0.13$^{1}$ \\
\midrule Mean, all walks & 0.00$^{11}$ & 0.09$^{1}$ & 0.10$^{2}$ & 0.13 & 0.11$^{3}$ & 0.12$^{2}$ \\
Beyond 3\,m, \% & \na & 15 & 5 & 9 & 10 & 13 \\
 
\bottomrule
\end{tabular}

\end{table*}

\paragraph{The GoPro walk.} A GoPro HERO13 Black was carried by hand for \gpMinutes{} minutes and \gpMetres\,m over two levels of a busy shopping mall (\cref{fig:gopro}). It records 4K video at \gpFrameHz\,Hz through its linear lens and its own IMU at \gpImuHz\,Hz; the system reads the video at $640{\times}360$. No calibration target was used: the intrinsics follow from the lens's nominal field of view, the camera's pose on the IMU is taken as the identity, and the IMU's noise is taken as a phone's. The camera's
electronic stabilization was on, which turns every image away from the IMU. We undo it in software from the orientations the camera stores with the video. Our run masks people out of tracking and mapping with a person detector (YOLOX); this switch is off in every scored run. The filter tracks the whole walk, and the map holds both levels and the stairs between them. There is no reference for this walk, so the result is qualitative.

\section{The depth model}\label{sec:s_ablation}\label{sec:s_model}
Two tests add to \cref{sec:exp_model}: whether the depth model explains the gap to ScaRF-SLAM's mapper, and which image the depth model is given.

\paragraph{ScaRF-SLAM's mapper with our depth model.} The ScaRF-SLAM row of \cref{tab:overview} runs its mapper with its own
depth model on the poses of one of our runs. Given our DA3-Giant checkpoint instead, on the same poses, it scores
\armScarfGiantRooms{} on the rooms and \armScarfGiantOri{} on ORI, against \armScarfOwnRooms{} and
\armScarfOwnOri{} with its own: the gap between the two mappers on the same poses is not the depth model.

\paragraph{The depth model and its input.} \Cref{tab:s_seq} gives the three depth models per sequence
(\cref{tab:model} gives the means). On a color camera the depth model is given the gray image
the filter tracks: on ORI with the giant model, replays of the same windows score \resInputGiantGrey{} with the
gray image against \resInputGiantColour{} with the color frame, the gray image ahead on every sequence; with the
base model the two inputs score alike (\resInputBaseGrey{} and \resInputBaseColour). The choice was made on the
development sequence R01, and we have no explanation for the difference between the models.

\section{Compute}\label{sec:s_compute}
\paragraph{How speed and memory were measured.} \Cref{tab:speed_memory} gives one measurement run per system on EuRoC, each in its released configuration, on the machine of our runs (one RTX~4090 with 24\,GB, a Xeon Gold 5320, 503\,GB of RAM). Memory is the peak over the run. Speed is the length of the footage over the processing time, given as frames per second of the camera's 20\,Hz. \method{} runs live, played at sensor rate, so it cannot be faster than 20, and its time includes loading the depth model.

\paragraph{Two GPU memories.} The GPU memory of \cref{tab:speed_memory} is what the card's driver reports for the whole process, the framework's cache included. \Cref{tab:model} reports a smaller number, the peak memory the depth model itself allocated (\resLiveGiantGpu\,GB for DA3-Giant, \resLiveBaseGpu\,GB for DA3-Base): the first says how much of a card the system occupies, the second how much the model needs.

\section{Limits}\label{sec:s_limits}
\emph{The scale is the filter's.} The anchor carries the filter's scale into the map and the correction keeps it.

\emph{The trajectory is the filter's.} It trails the best VIO systems (\cref{sec:s_results}), drifts on Hilti-Oxford's construction site while the rig stands before a blank wall (\cref{sec:s_hilti}), fails on three of
the phone recordings (\cref{sec:s_devices}), and on one Oxford Spires sequence loop closure makes it worse (\cref{sec:s_spires}). Given the reference's poses in place of the filter's, the mapper scores \armRefRoomsCorr{} on the rooms and \armRefOriCorr{} on ORI, against \armRecRoomsCorr{} and \armRecOriCorr{} with the filter's poses.

\emph{Range.} Depth beyond 20\,m is dropped, so outdoors the far sides of large open spaces are not mapped.

\FloatBarrier

\end{document}